\documentclass[sigconf]{acmart}

\acmConference[ICSE 2022]{The 44th International Conference on Software Engineering}{May 21-29, 2022}{Pittsburgh, PA, USA}

\usepackage{enumitem}
\usepackage{enumerate}
\usepackage{float}
\usepackage{hyperref}       % hyperlinks
\usepackage{url}            % simple URL typesetting
\usepackage{booktabs}       % professional-quality tables
\usepackage{amsfonts}       % blackboard math symbols
\usepackage{nicefrac}       % compact symbols for 1/2, etc.
\usepackage{microtype}      % microtypography
\usepackage{soul}
\usepackage{bm}
\usepackage{bbding}
\usepackage{lineno}
\usepackage{multirow}
\usepackage{amsmath}
\usepackage{color}
\usepackage{pifont}
\usepackage{mathrsfs}

\usepackage{amsmath,amssymb,amsfonts}
\usepackage{algorithmic}
\usepackage{subfigure}
\usepackage{graphicx}
\usepackage{textcomp}
\usepackage{xcolor}
\graphicspath{ {images/} }

\def\BibTeX{{\rm B\kern-.05em{\sc i\kern-.025em b}\kern-.08em
    T\kern-.1667em\lower.7ex\hbox{E}\kern-.125emX}}

\begin{document}

\title{NeuronFair: Interpretable White-Box Fairness Testing through Biased Neuron Identification}

\author{Haibin Zheng}
\orcid{0000-0002-8997-5343}
\affiliation{%
  \institution{\small{Zhejiang University of Technology}}
  \country{}
}
\email{haibinzheng320@gmail.com}

\author{Zhiqing Chen}
\affiliation{%
  \institution{\small{Zhejiang University of Technology}}
  \country{}
}
\email{201706060201@zjut.edu.cn}

\author{Tianyu Du}
\affiliation{%
  \institution{\small{Zhejiang University}}
  \country{}
%   \city{Hangzhou}
%   \country{China}
%   \postcode{310007}
}
\email{zjradty@zju.edu.cn}

\author{Xuhong Zhang}
\affiliation{%
  \institution{\small{Zhejiang University}}
  \country{}
%   \city{Hangzhou}
%   \country{China}
%   \postcode{310007}
}
\email{zhangxuhong@zju.edu.cn}

\author{Yao Cheng}
\affiliation{%
  \institution{\small{Huawei International Pte. Ltd.}}
  \country{}
%   \streetaddress{11 North Buona Vista Drive #17-08}
%   \city{The Metropolis Tower 2}
%   \country{Singapore}
%   \postcode{138589}
}
\email{chengyao101@huawei.com}

\author{Shouling Ji}
\affiliation{%
  \institution{\small{Zhejiang University}}
  \country{}
%   \city{Hangzhou}
%   \country{China}
%   \postcode{310007}
}
\email{sji@zju.edu.cn}

\author{Jingyi Wang}
\affiliation{%
  \institution{\small{Zhejiang University}}
  \country{}
%   \city{Hangzhou}
%   \country{China}
%   \postcode{310007}
}
\email{wangjyee@zju.edu.cn}

\author{Yue Yu}
\affiliation{%
  \institution{\small{National University of Defense Technology}}
  \country{}
%   \city{Hefei}
%   \country{China}
%   \postcode{230000}
}
\email{yuyue@nudt.edu.cn}

\author{Jinyin Chen}
%\authornotemark[1]
\authornote{Corresponding author.}
\affiliation{%
  \institution{\small{Zhejiang University of Technology}}
  \country{}
%   \city{Hangzhou}
%   \country{China}
%   \postcode{310023}
}
\email{chenjinyin@zjut.edu.cn}

 \renewcommand{\shortauthors}{Zheng, et al.}

\begin{abstract}
Deep neural networks (DNNs) have demonstrated their outperformance in various domains.
However, it raises a social concern whether DNNs can produce reliable and fair decisions especially when they are applied to sensitive domains involving valuable resource allocation,
such as education, loan, and employment.
It is crucial to conduct fairness testing before DNNs are reliably deployed to such sensitive domains,
i.e., generating as many instances as possible to uncover fairness violations.
However, the existing testing methods are still
limited from three aspects: interpretability, performance, and generalizability.
To overcome the challenges,
we propose \emph{NeuronFair},
a new DNN fairness testing framework that differs from previous work in several key aspects:
(1)~\emph{interpretable} - it quantitatively interprets DNNs' fairness violations for the biased decision;
(2)~\emph{effective} - it uses the interpretation results to guide the generation of more diverse instances in less time;
(3)~\emph{generic} - it can handle both structured and unstructured data.
Extensive evaluations across 7 datasets and the corresponding DNNs demonstrate NeuronFair's superior performance.
For instance, on structured datasets,
% it generates much more instances ($\sim\!\times$7.32) in much less time (1/2 to 1/20) compared with the state-of-the-art methods.
it generates much more instances ($\sim\!\times$5.84) and saves more time (with an average speedup of 534.56\%) compared with the state-of-the-art methods.
Besides,
the instances of NeuronFair can also be leveraged to improve the fairness of the biased DNNs,
which helps build more fair and trustworthy deep learning systems.
The code of NeuronFair is open-sourced at
\textit{\url{https://github.com/haibinzheng/NeuronFair}}.
% \textit{\url{https://anonymous.4open.science/r/NeuronFair-5652/}}.
\end{abstract}






%\linenumbers
\maketitle

\section{Introduction}
% P1:引出针对深度学习中的公平性测试很重要
% S1：深度学习很重要的应用。
Deep neural networks (DNNs)~\cite{LeCun2015Deep} have been increasingly adopted in many fields,
including computer vision~\cite{Badar2020Application},
natural language processing~\cite{devlin2018bert},
software engineering~\cite{devanbu2020deep,Chen2020Software,Sedaghatbaf2021Automated,Mirabella2021Deep,Huang2021Robustness,Li2021DeepPayload}, etc.
% S2：然而深度学习存在很多问题，其中之一就是决策的偏见。
However, one of the crucial factors hindering DNNs from further serving applications with social impact is the unintended individual discrimination~\cite{Meng2021Measuring,Ramadan2018Model,Zhang2021Ignorance}.
Individual discrimination exists when a given instance different from another only in sensitive attributes (e.g., gender, race, etc.) but receives a different prediction outcome from a given DNN~\cite{Aggarwal2019Black}.
% S3-S4：（用一个例子）解释一下什么是偏见。尤其在其他一些敏感应用中就更需要被研究。
Taking gender discrimination in salary prediction as an example,
for two identical instances except for the gender attribute,
male's annual income predicted by the DNN is often higher than female's~\cite{Kohavi1996Scaling}.
Thus, it is of great importance for stakeholders
to uncover fairness violations and then
to reduce DNNs' discrimination
so as to responsibly deploy fair and trustworthy deep learning systems in many sensitive scenarios~\cite{Huang2018DeepCrime,Gaur2020Semi,Mai2019Deep,Buolamwini2018Gender}.

% P2: 分类介绍现有方法及其存在的缺陷和难点，引出我们的design goal
% S1：为了实现公平，所以目前提出了很多测试方法。
Much effort has been put into uncovering fairness violations~\cite{Farahani2021OnAdaptive,Clegg2019Simulating,German2018WasMyContribution,Holstein2018Avoiding,Melton2018Onfairness,Verma2018Fairness,Biswas2021FairPreprocessing,Biswas2020DoTheMachine,Brun2018Software}.
The most common method is fairness testing~\cite{Dwork2012Fairness,Aggarwal2019Black,Black2020FlipTest,Galhotra2017Fairness,Aggarwal2018Automated,Udeshi2018Automated,Zhang2020White,Zhang2021Efficient},
which solves this problem by generating as many instances as possible.
% S2-S3：首先，针对传统机器学习的，有哪些方法。但是应用到深度学习中，一些方法存在这种问题，一些方法存在那种问题。
Initially, fairness testing is designed to uncover and reduce the discrimination in traditional machine learning (ML) with low-dimensional linear models.
However, such methods are suffering from several problems.
First, most of them (e.g.,
FairAware~\cite{Dwork2012Fairness},
BlackFT~\cite{Aggarwal2019Black},
and FlipTest~\cite{Black2020FlipTest})
cannot handle DNNs with high-dimensional nonlinear structures.
Then, though some of them (e.g.,
Themis~\cite{Galhotra2017Fairness},
SymbGen~\cite{Aggarwal2018Automated},
and Aequitas~\cite{Udeshi2018Automated})
can be applied to test DNNs,
they are still challenged by the high time cost and numerous duplicate instances.
Recently, several methods have been specifically developed for DNNs,
such as ADF~\cite{Zhang2020White} and EIDIG~\cite{Zhang2021Efficient}, etc.
These methods make progress in effectiveness and efficiency through gradient guidance,
but they still suffer from the following problems.

First, these methods can hardly be generalized to unstructured data.
As we know, DNNs are originally designed to process unstructured data (e.g., image, text, speech, etc.),
but almost no existing fairness testing method can be applied to these data.
It is mainly because these methods
cannot determine which features are related to sensitive attributes,
and cannot implement appropriate modifications to these features,
e.g., how to determine pixels related to gender attribute in face images,
and how to modify these pixel values to change gender~\cite{Wang2020TowardsVisual}.
However, even a seemingly simple task such as face detection~\cite{Klare2012Face} is subject to extreme amounts of fairness violations.
It is especially concerning since these facial systems are often not deployed in isolation but rather as part of the surveillance or criminal detection pipeline~\cite{Amini2019Uncovering}.
Therefore,
these testing methods still cannot serve DNNs widely until we solve the problem of data generalization.

Second, the generation effectiveness of these methods is challenged by gradient vanishing.
They leverage the gradient-guided strategy to improve generation efficiency,
but the gradient may vanish and cause instance generation to fail.
Additionally, when the gradient is small,
the generated instances are highly similar.
However, the purpose of fairness testing is to generate not only the numerous instances,
but also the diverse instances.

Third, almost all existing methods hardly provide interpretability.
They only focus on generating numerous instances,
but cannot interpret how the biased decisions occurred.
DNNs' decision results are determined by neuron activation,
then we try to study these neurons that cause biased decisions.
We find that the instances generated by existing testing methods will miss the coverage of these neurons that cause biased decisions (refer to the experiment result in Fig.~\ref{fig:NeuronCoverage}).
More seriously, we cannot even know which neurons related to biased decisions have been missed for testing
when there is a lack of interpretability.
Therefore,
we need an interpretable testing method so as to
interpret DNNs' biased decisions and evaluate instances' utility for uncovering fairness violations.
Based on these,
the interpretation results can guide us to design effective testing to uncover more discrimination.
In summary, the current fairness testing challenges lie in
the lack of data generalization,
generation effectiveness,
and discrimination interpretation.

% P3-P5：针对上述三个难点，介绍我们的设计目标，我们希望它有哪些优势来解决上述难点。
% S1：我们的目标是设计一种怎样的测试方法。
To overcome the above challenges,
our design goals are as follows:
1)~we intend to uncover and quantitatively interpret DNNs' discrimination;
2)~then, we plan to apply this interpretation results to guide fairness testing;
3)~furthermore, we want to generalize our testing method to unstructured data.
% S2-：为了实现定量的偏见解释，有哪些措施，为什么这么设计，带来哪些益处。
Due to the decision results of DNNs are determined by the nonlinear combination of each neuron's activation state,
thus we imagine whether the biased decisions are caused by some neurons.
Then, we try to observe the neuron activation state in DNNs' hidden layers through feeding instance pair,
which is two identical instances except for the sensitive attribute.
Surprisingly, we find that the activation state follows such a pattern,
i.e.,
neurons with drastically varying activation values are overlapping for different instance pairs.
We observe that DNNs' discrimination is reduced when these overlapped neurons are zeroed out.
Therefore, we conclude that these neurons cause the DNNs' discrimination.
Then, we intend to quantitatively interpret DNNs' discrimination by computing the neuron activation difference (ActDiff) values.

% P4: 为了实现优化梯度求导，有哪些措施，为什么这么设计，带来哪些益处。
According to the interpretation results,
we further design a testing method, NeuronFair, to optimize gradient guidance.
First, we determine the main neurons that cause discrimination,
called biased neurons.
Then, we search for discriminatory instances with the optimization object of increasing the ActDiff values of biased neurons.
Because the optimization from the biased neuron shortens the derivation path,
it reduces the probability of the gradient vanishing
and time cost.
Moreover, we can produce more diverse instances through the dynamic combination of biased neurons.
All in all, we leverage the interpretation results to optimize gradient guidance,
which is beneficial to the generation effectiveness.

% P5：为了实现非结构数据的属性操作，有哪些措施，为什么这么设计，带来哪些益处。
We leverage adversarial attacks~\cite{goodfellow2014explaining,Kurakin2017Adversarial,chen2020mag} to determine which features are related to sensitive attributes,
and make appropriate modifications to these features.
The adversarial attack is originally to test the DNNs' security,
e.g., slight modifications to some image pixels will cause the predicted label to flip~\cite{Brendel2018Boundary,chen2017zoo,Dong2018Boosting,Tabacof2016Exploring}.
Taking the gender attribute of face image as an example,
we consider training a classifier with `male' and `female' as labels,
then adding the perturbation to the face image until its predicted gender label flips.
Based on this generalization framework,
we can modify the sensitive attributes of any data,
thereby generalizing NeuronFair to any data type.

% 【贡献】
In summary,
we first implement to quantitatively interpret the discrimination using neuron-based analysis;
then, we leverage the interpretation results to optimize the instance generation;
finally, we design a generalization framework for sensitive attribute modification.
The main contributions are as follows.

\begin{itemize}[leftmargin=9pt]
  \item Through the neuron activation analysis,
        we quantitatively interpret DNNs' discrimination,
        which provides a new perspective for measuring discrimination and guides DNNs' fairness testing.
  \item Based on the interpretation results,
        we design a novel method for DNNs' discriminatory instance generation, NeuronFair,
        which significantly outperforms previous works in terms of effectiveness.
  \item Inspired by adversarial attacks,
        we design a generalization framework to modify sensitive attributes of unstructured data,
        which generalizes NeuronFair to unstructured data.
  \item We publish our NeuronFair as a self-contained open-source toolkit online.
\end{itemize}

% % 【文章的整体结构】

\begin{figure}[t]
  \centering
  \includegraphics[width=\linewidth]{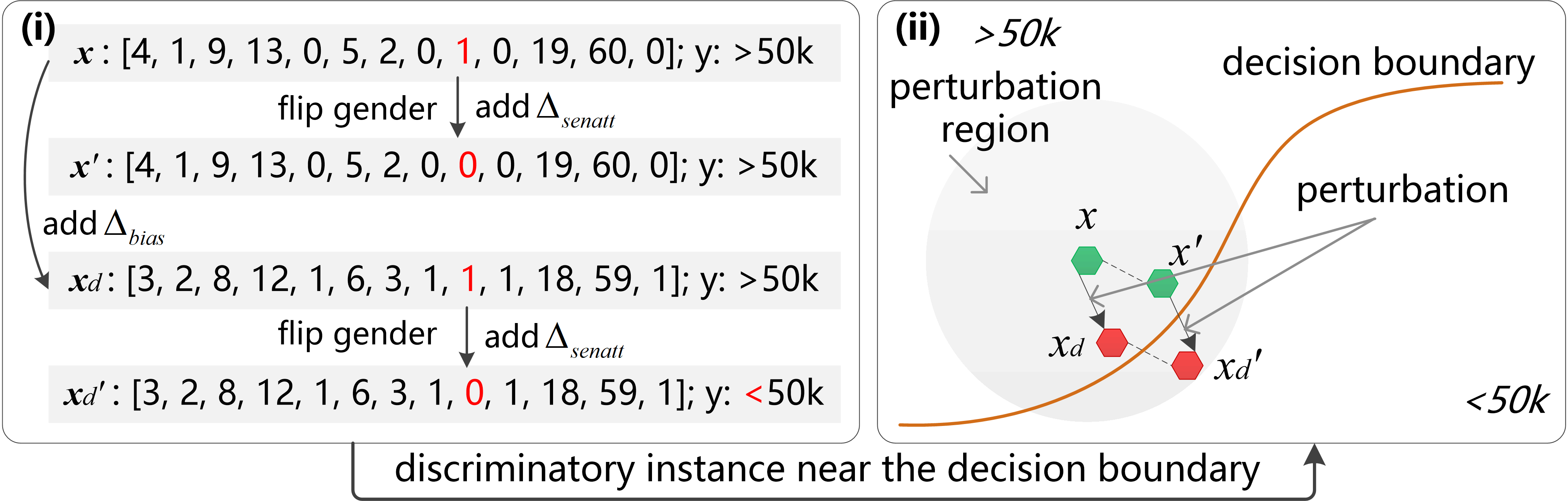}
  \caption{Illustration of discriminatory instance generation on Adult dataset~\cite{Kohavi1996Scaling}.
            \textbf{(i)}
            The discriminatory instance generation process.
            The normal instance pair is $<\!x, x'\!>$ and the discriminatory instance pair is $<\!x_{d}, x_{d}'\!>$.
            $x'$=$x$+$\Delta_{senatt}$, $x_{d}$=$x$+$\Delta_{bias}$, $x_{d}'$=$x$+$\Delta_{senatt}$+$\Delta_{bias}$,
            where
            $\Delta_{bias}$ is the bias perturbation,
            $\Delta_{senatt}$ is the perturbation added to the gender attribute to flip gender.
            \textbf{(ii)}
            % The discriminatory instance exists near the decision boundary.
            Discrimination exists when the instance's predicted label changes as the gender attribute is flipped,
            i.e., the instance crosses the decision boundary.
  }
  \label{fig:schematicBoundaryBased}
  \vspace{-0.4cm}
\end{figure}

\section{Background\label{Background}}
To better understand the problem we are tackling and the methodology we propose in later sections,
we first introduce DNN, data form, individual discrimination, and our problem definition.

\emph{\textbf{DNN}}.
A DNN can be represented as $f(x;\Theta):\mathcal{X}\to{\mathcal{Y}}$,
including an input layer, several hidden layers, and an output layer.
Two popular architectures of DNNs are fully connected network (FCN)
and convolutional neural network (CNN).
For a FCN,
we denote the activation output of each neuron in the hidden layer as:
$f_{l}^{k}(x; \Theta)$,
where $\Theta$ is the weights,
$l\in\{1,2,...,N_{l}\}$,
$N_{l}$ is the number of neural layers,
$k\in\{1,2,...,N_{l}^{k}\}$,
$N_{l}^{k}$ is the number of neurons in the $l$-th layer.
For a CNN,
we flatten the output of the convolutional layer for the calculation of neuron activation.
The loss function of DNNs is defined as follows:
\vspace{-0.05cm}
\begin{equation}\label{equ:mulCE}
 J(x,y;\Theta) = -\frac{1}{N} \left[
    \sum\nolimits_{i=0}\nolimits^{N-1}
    \sum\nolimits_{j=0}\nolimits^{M-1}
        \left({y_{i,j} \!\times\! \log({\hat{y}_{i,j}})}\right)\right]
\vspace{-0.05cm}
\end{equation}
where $N$ is the number of instances,
$M$ is the number of classes,
$y_{i}$ is the ground-truth of $x_{i}$,
$\hat{y_{i}} = f(x_{i};\Theta)$ is the predicted probability,
$\log(\cdot)$ is a logarithmic function.

\emph{\textbf{Data Form}}. Denote $X=\{x_{i}\}$, $Y=\{y_{i}\}$ as a normal dataset,
and its instance pairs by $<\!X, X'\!>=\{<\!x_{i},x_{i}'\!>\}$, $i\in\{0,1,...,N-1\}$.
For an instance, we denote its attributes by $A=\{a_{i}\}$, $i\in\{0,1,...,N_{a}-1\}$,
where $A_s\subset{A}$ is a set of sensitive attributes,
and $A_{ns} =\{a_{i}^{ns}|a_{i}^{ns}\in{A}, {\rm and} ~a_{i}^{ns}\notin{A_{s}}\}$ is a set of non-sensitive attributes.
% 典型的受保护属性如性别、种族、年龄等，通常根据具体的敏感场景预先给定。
Note that sensitive attributes (e.g., gender, race, age, etc.) are usually given in advance according to specific sensitive scenes.

\emph{\textbf{Individual Discrimination}}.
As stated in previous work~\cite{Dwork2012Fairness,Kusner2017Counterfactual,Aggarwal2019Black},
individual discrimination exists when two valid inputs which differ only in the sensitive attributes but receive a different prediction result from a given DNN,
as shown in Fig.~\ref{fig:schematicBoundaryBased}.
Such two valid inputs are called individual discriminatory instances (IDIs).

\emph{Definition 1: IDI determination.}
We denote $<\!X_{d}, X_{d}'\!>=\{<\!x_{d,i},x_{d,i}'\!>\}$
as a set of IDI pairs,
which satisfies:
\vspace{-0.2cm}
\begin{equation}\label{equ:def1}
    \begin{array}{c}
      f(x_{d,i};\Theta) \neq f(x'_{d,i};\Theta) \\
      {\rm s.t.}~~x_{d,i}{\rm [A_{s}]}\neq x_{d,i}'{\rm [A_{s}]}, ~x_{d,i}{\rm [A_{ns}]} = x_{d,i}'{\rm [A_{ns}]}
    \end{array}
\end{equation}
where $i\in\{0,1,...,N_{d}-1\}$,
$x_{d,i}{\rm [A_{s}]}$ represents the value of $x_{d,i}$ with respect to attribute $A_{s}$.
Note that our instances are generative (e.g., maybe the age of a generated instance is 150 years old on Adult dataset),
thus we need to clip their attribute values that do not exist in the input domain $\mathbb{I}$.

\textit{\textbf{Problem Definition}}.
A DNN which suffers from individual discrimination may produce biased decisions when an IDI is presented as input.
Below are the three goals that we want to achieve through the devised fairness testing technique.
First, observe the activation state of neurons,
find the correlation between neuron activation pattern and biased decision,
and interpret the reason for discrimination.
Then, based on the interpretation results,
we generate IDIs with maximizing DNN's discrimination as the optimization object.
In the generation process,
we consider not only the generation quantity, but also the diversity.
Finally, we conduct fairness testing on unstructured datasets.

\section{NeuronFair\label{Methodology}}
% framework
\begin{figure}[t]
\centering
    \includegraphics[width=\linewidth]{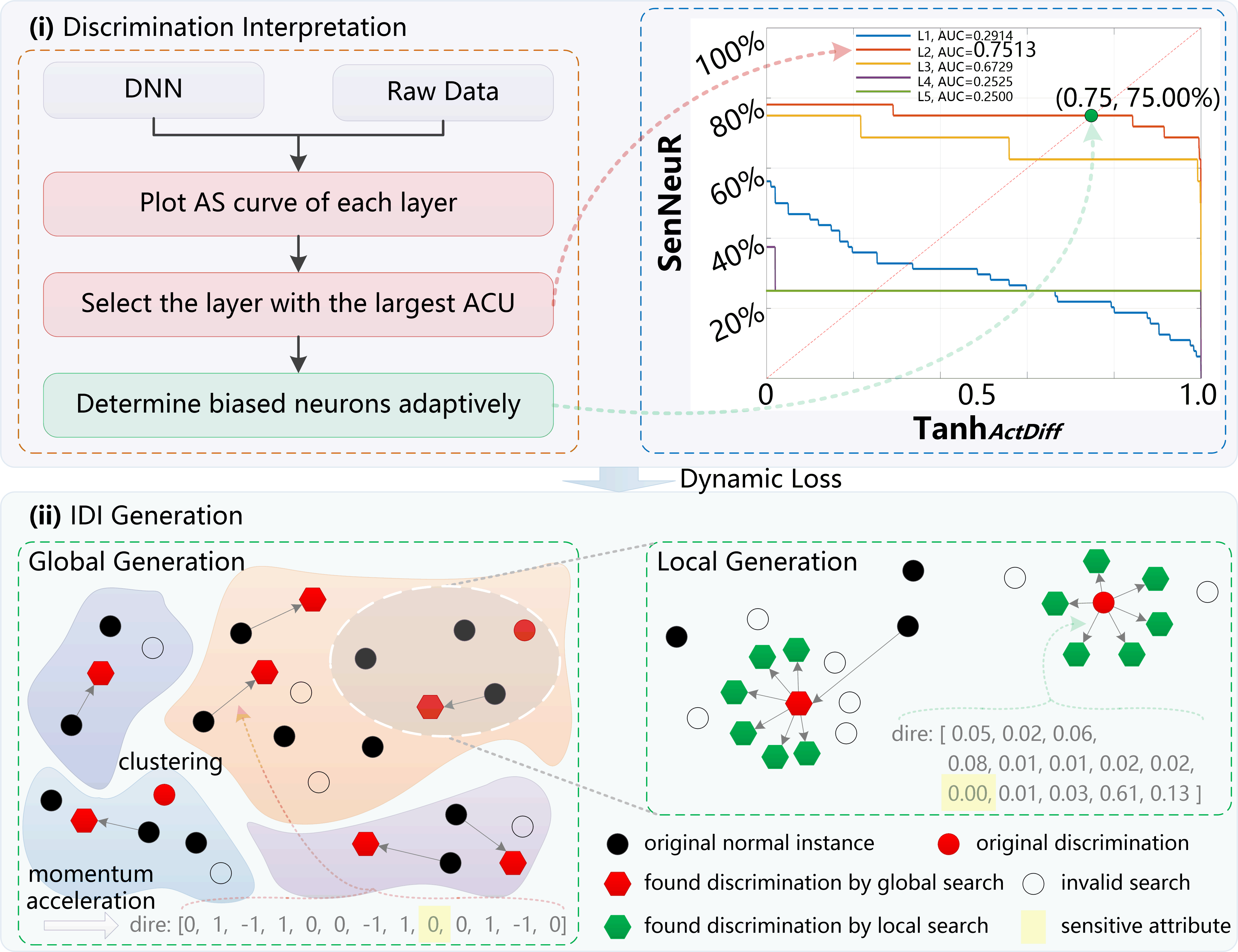}
  \caption{An overview of NeuronFair.%, where the DNN and dataset are the same as Fig.~\ref{fig:schematic}.
        }
  \label{fig:framework}
  \vspace{-0.4cm}
\end{figure}

% 概述NeuronFair
An overview of NeuronFair is presented in Fig.~\ref{fig:framework}.
NeuronFair has two parts, i.e., discrimination interpretation and IDI generation based on interpretation results.
% 解释期间：
% 我们首先通过分析神经元解释DNN的歧视，
% 然后根据解释得到的歧视原因设计先验的歧视度量指标，
% 最后根据歧视度量自动发现偏见神经元，服务于后续的歧视测试。
During the discrimination interpretation,
we first interpret why discrimination exists through neuron-based analysis.
Then, we design a discrimination metric based on the interpretation result, i.e., AUC value,
as shown in Fig.~\ref{fig:framework} (i).
AUC is the area under AS curve,
where the AS curve records the percentage of neurons above the ActDiff threshold.
Finally, we leverage the AS curve to adaptively identify biased neurons,
which serves for IDI generation.
% 生成期间：
% 包括全局生成和局部生成，前者保证多样性，后者保证数量。
% 使用偏见神经元作为指导：加快了搜索速度、降低了搜索过程的梯度消失概率、增加了搜索结果的动态性（多样性）
During the IDI generation,
we employ the biased neurons to perform global and local generations.
The global phase guarantees the diversity of the generated instances,
and the local phase guarantees the quantity,
as shown in Fig.~\ref{fig:framework} (ii).
On the one hand,
the global generation uses the normal instance as a seed and
stops if an IDI is generated or it times out.
On the other hand,
the generated IDIs are adopted as seeds of local generation,
leading to generate as many IDIs as possible near the seeds.
Besides, we implement dynamic combinations of biased neurons to increase diversity,
and use the momentum strategy to accelerate IDI generation.
% 方法部分的组织
In the following,
we first quantitatively interpret DNNs' discrimination,
then present details of IDI generation based on interpretation results,
and finally generalize NeuronFair to unstructured data.

\subsection{Quantitative Discrimination Interpretation}
% % 在测试之前，解释每一层中存在的歧视是十分必要的。
% % 1 可以帮助直观的了DNN中歧视存在的位置。
% % 2 可以帮助我们从歧视最严重的的位置开始搜索歧视样本，改善效果和效率。

First, we draw AS curve and compute AUC value to measure DNNs' discrimination.
Then, based on the measurement results, we identify the key neurons that cause unfair decisions as biased neurons.

\subsubsection{Discrimination Measurement}

The ActDiff is calculated as follows:
\vspace{-0.1cm}
\begin{equation}\label{equ:ActDiff}
    z_{l}^{k} = \frac{1}{N}\sum\nolimits_{i=0}\nolimits^{N\!-\!1}[{\rm abs}(f_{l}^{k}(x_{i};\Theta)-f_{l}^{k}(x_{i}';\Theta))]
\vspace{-0.1cm}
\end{equation}
where $z_{l}^{k}$ is the ActDiff of the $k$-th neuron in the $l$-th layer,
$l\in\{1,2,...,N_{l}\}$,
$N_{l}$ is the layer number,
$N$ is the number of normal instance pairs $<\!X,X'\!>=\{<\!x_{i},x_{i}'\!>\}$,
$i\in\{0,1,2,...,N\!-\!1\}$,
$\rm abs(\cdot)$ returns an absolute value,
$f_{l}^{k}(x;\Theta)$ returns the activation output of the $k$-th neuron in the $l$-th layer,
$\Theta$ represents the model weights.

\begin{figure}[t]
  \centering
    \includegraphics[width=\linewidth]{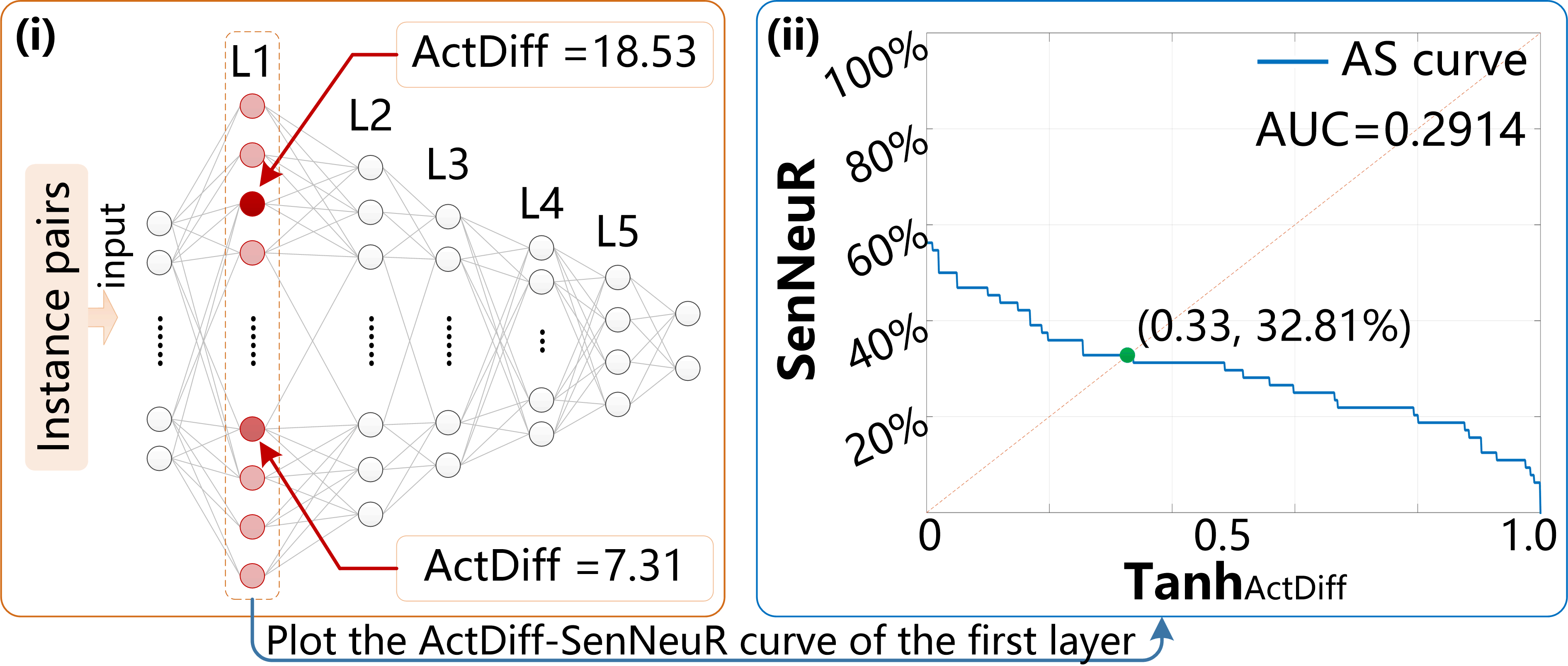}
  \caption{Illustration of the neuron-based discrimination interpretation.
        Dataset: Adult~\cite{Kohavi1996Scaling};
        dimension: 13;
        % instance number: 32,561;
        FCN-based DNN structure: [13, 64, 32, 16, 8, 4, 1].
  }
  \label{fig:schematicNeuronBased}
  \vspace{-0.2cm}
\end{figure}

Based on Eq.~(\ref{equ:ActDiff}),
we plot AS curve and compute AUC value.
We first compute each neuron's ActDiff and normalize it by hyperbolic tangent function ${\rm Tanh}(\cdot)$,
as shown in Fig.\ref{fig:schematicNeuronBased} (i).
``L1'' means the $1$-st hidden layer of a DNN with 64 neurons.
Then, we set several ActDiff thresholds at equal intervals,
count the neuron percentages above the ActDiff thresholds,
and record them as sensitive neuron rate (SenNeuR).
Finally, we plot AS curve according to the SenNeuR under different ActDiff thresholds,
and then compute the area under AS curve as AUC value,
as shown in Fig.\ref{fig:schematicNeuronBased} (ii),
where the x-axis is the ActDiff value normalized by Tanh function,
the y-axis is SenNeuR.
Repeat such an operation for each layer,
we can intuitively observe the discrimination in each layer
and find the most biased layer with the largest AUC value.
As shown in Fig.~\ref{fig:framework} (i),
the $2$-nd layer `L2' is selected as the most biased layer with AUC=0.7513.

More specific operations on AS curve drawing and AUC calculation are shown in \textbf{Algorithm 1}.
First, we compute the average ActDiff values of each neuron $z_{l}^{k}$ at line 1.
In the loop from lines 7 to 9, for each neural layer,
we get SenNeuR for plotting the AS curve.
Then, we compute AUC value by integration at line 11.
% Finally, the layer with the largest AUC value is selected for adaptive biased neuron search.

% Algorithm 1
{
\renewcommand\arraystretch{0.85}
\begin{table}[t]
  \centering
  \begin{tabular}{lp{0.86\columnwidth}}
  \toprule
    \multicolumn{2}{l} {\textbf{Algorithm 1}:
    % A priori discrimination measurement, i.e., AS curve and AUC value.
    AS curve drawing and AUC calculation.
    } \\ \hline
      &
    \textbf{Input}: The activation output $f_{l}^{k}(x;\Theta)$,
                    ActDiff threshold interval $step_{-}interval$ = 0.005,
                    instance pairs $<\!X,X'\!>$.
    \\
      &
    \textbf{Output}: AS curve and AUC value of each layer.
    \\ \cline{2-2}
    1 &
    Calculate the average ActDiff  of each neuron:
    \\
        &
    \quad $z_{l}^{k} = \frac{1}{N}\sum_{i=0}^{N\!-\!1}[{\rm abs}(f_{l}^{k}(x_{i};\Theta)-f_{l}^{k}(x_{i}';\Theta))]$
    \\
    2 &
    \textbf{For} $l=1:N_{l}$
    \\
    3 &
    \quad $z_{l} = {\rm{Tanh}}(z_{l})$
    \\
    4 &
    \quad $max_{-}z = {\rm max}(z_{l})$
    \\
    5 &
    \quad $x_{tmp} = 0:step_{-}interval:max_{-}z$
    \\
    6 &
    \quad $y_{tmp} = [~]$
    \\
    7 &
    \quad \textbf{For} $count=1:{\rm length}(x_{tmp})$
    \\
    8 &
    \qquad $y_{tmp} = [y_{tmp}, {\rm length}({\rm find}(z_{l} > x_{tmp}[count]))]$
    \\
    9 &
    \quad \textbf{End For}
    \\
    10 &
    \quad $y_{tmp} = y_{tmp}/{\rm length}(z_{l})\times 100\%$
    \\
    11 &
    \quad $area = \sum_{count = 1}(y_{tmp}[count]\times{step_{-}interval})$
    \\
    12 &
    \quad Plot the AS curve based on ($x_{tmp}$, $y_{tmp}$).
    \\
       &
    \quad Save $area$ as the AUC of the $l$-th layer.
    \\
    13 &
    \textbf{End For}
    \\
  \bottomrule
  \end{tabular}
  \vspace{-0.3cm}
\end{table}
}

\subsubsection{Biased Neuron Identification}
The most biased layer is selected for adaptive biased neuron identification.
A neuron with a large $z_{l}^{k}$ value demonstrates that it responds violently to the modification of sensitive attributes,
thus it carries more discrimination.
We define biased neuron as follows.

\emph{Definition 2: Biased neuron.}
For a given discrimination threshold $T_{d}$ of the most biased layer,
the biased neurons satisfy the condition $z^{k}>T_{d}$.
$z^{k}$ is the average ActDiff  normalized by ${\rm Tanh}(\cdot)$ of the $k$-th neuron in the most biased layer,
$k\in\{1,2,...,N^{k}\}$,
$T_{d}\in(0,1)$.

Based on the \emph{Definition 2},
we know that once $T_{d}$ is determined,
biased neurons can be found.
Here we give a strategy for adaptively determining $T_{d}$.
We draw a line $y=x$ that intersects the AS curve.
The x-axis's value of this intersection is $T_{d}$.
As shown in Fig.~\ref{fig:schematicNeuronBased} (ii),
the intersection is the point (0.33, 32.81\%) and $T_{d}$=0.33,
After determining $T_{d}$,
we record these biased neurons and save their position $p$,
where $p$ is a vector with $N^{k}$ elements.

\subsection{Interpretation-based IDI Generation}
NeuronFair generates IDIs in two phases,
i.e., a global generation phase and a local generation phase.
% Both of them adopt dynamic loss function and momentum acceleration.
The global phase aims to acquire diverse IDIs. % that increase the ActDiff of the biased neurons.
The IDIs' diversity in the global phase is crucial since these instances serve as seeds for the local phase.
Instead, to guarantee the IDIs' quantity,
the local phase aims to search for as many IDIs as possible near the seeds.

% Algorithm 3->2
{
\renewcommand\arraystretch{0.85}
\begin{table}[t]
  \centering
  \begin{tabular}{lp{0.86\columnwidth}}
  \toprule
    \multicolumn{2}{l} {\textbf{Algorithm 2}: Global generation guided by biased neurons.} \\ \hline
      &
    \textbf{Input}: Normal instance $X=\{x_{i}\}$,
                    initial set $\Omega_{g}=\varnothing$,
                    $X_{c}$ = KMeans($X$, $N_{c}$), $c\in\{1,2,...,N_{c}\}$,
                    the number of seeds for global generation $num_g$,
                    the maximum number of iterations for each seed $max_{-}iter_{g}$,
                    the perturbation size of each iteration $step_{-}size_{g}$,
                    the decay rate of momentum $\mu_{g}$,
                    the step size for random disturbance $r_{-}step_{g}$.
                    % Here we default that $num_{g}/N_{c}<\min ({\rm length} (X_{c}) )$.
                    %\cy{we default??}
    \\
      &
    \textbf{Output}: A set of IDI pairs found globally $\Omega_{g}$.
    \\ \cline{2-2}
    1 &
    \textbf{For} $i=0:{\rm INT}(num_g/N_{c})\!-\!1$
    \\
    2 &
    \quad \textbf{For} $c=1:N_{c}$
    \\
    3 &
    \qquad Select seed $x$ from $X_{c}$, $g_{0}=g'_{0}=0$.
    \\
    4 &
    \qquad \textbf{For} $t=0:max_{-}iter_{g}$
    \\
    5 &
    \qquad \quad \textbf{If} ( mod($step_{-}size_{g}$, $r_{-}step_{g}$) == 0 ) \textbf{Then}
    \\
    6 &
    \qquad \qquad $r=Rand_{(0,1)}(p_{r})$
    \\
    7 &
    \qquad \quad \textbf{End If}
    \\
    8 &
    \qquad \quad Create {\footnotesize $<\!x,x'\!>$ s.t. $x{\rm [A_{s}]}\!\neq x'{\rm [A_{s}]}$, $x{\rm [A_{ns}]} \!= x'{\rm [A_{ns}]}$.}
    \\
    9 &
    \qquad \quad \textbf{If} ($f(x;\Theta)\neq f(x';\Theta)$) \textbf{Then}
    \\
    10 &
    \qquad \qquad $\Omega_{g}= \Omega_{g}\cup{<\!x,x'\!>}$
    \\
    11 &
    \qquad \qquad \textbf{break}
    \\
    12  &
    \qquad \quad \textbf{End If}
    \\
    13 &
    \qquad \quad $g_{t+1}=\mu_{g} \!\times\! g_{t} + \nabla_{x}J_{dl}(x;\Theta)$
    \\
    14 &
    \qquad \quad $g'_{t+1}=\mu_{g} \!\times\! g'_{t} + \nabla_{x'}J_{dl}(x';\Theta)$
    \\
    15 &
    \qquad \quad $dire = {\rm sign} (g_{t+1} \!+ g'_{t+1})$
    \\
    16 &
    \qquad \quad $dire{\rm [A_{s}]}=0$
    \\
    17 &
    \qquad \quad $x = x+ dire \!\times\! step_{-}size_{g}$
    \\
    18 &
    \qquad \quad $x = {\rm Clip}(x,\mathbb{I})$
    \\
    19 &
    \qquad\textbf{End For}
    \\
    20 &
    \quad \textbf{End For}
    \\
    21 &
    \textbf{End For}
    \\
  \bottomrule
  \end{tabular}
  \vspace{-0.3cm}
\end{table}
}

% Algorithm 4->3
{
\renewcommand\arraystretch{0.85}
\begin{table}[t]
  \centering
  \begin{tabular}{lp{0.86\columnwidth}}
  \toprule
    \multicolumn{2}{l} {\textbf{Algorithm 3}: Local generation guided by biased neurons.} \\ \hline
      &
    \textbf{Input}: IDI pairs $\Omega_{g}=\{<\!x_{d,i},x_{d,i}'\!>\}$,
                    $i=\{0,1,...,N_{g}\!-\!1\}$,
                    initial set $\Omega_{l}=\varnothing$,
                    the maximum number of iterations for each seed $max_{-}iter_{l}$,
                    the perturbation size of each iteration $step_{-}size_{l}$,
                    the decay rate of momentum $\mu_{l}$,
                    step size for random disturbance $r_{-}step_{l}$.
    \\
      &
    \textbf{Output}: A set of IDI pairs found locally $\Omega_{l}$.
    \\ \cline{2-2}
    1 &
    \textbf{For} $i=0:N_{g}\!-\!1$
    \\
    2 &
    \quad Select seed $<\!x,x'\!>$ from $\Omega_{g}$, $g_{0}=g'_{0}=0$.
    \\
    3 &
    \quad \textbf{For} $t=0:max_{-}iter_{l}$
    \\
    4 &
    \qquad \textbf{If} ( mod($step_{-}size_{l}$, $r_{-}step_{l}$) == 0 ) \textbf{Then}
    \\
    5 &
    \qquad \quad $r=Rand_{(0,1)}(p_{r})$
    \\
    6 &
    \qquad \textbf{End If}
    \\
    7 &
    \qquad $g_{t+1}=\mu_{l} \!\times\! g_{t} + \nabla_{x}J_{dl}(x;\Theta)$
    \\
    8 &
    \qquad $g'_{t+1}=\mu_{l} \!\times\! g'_{t} + \nabla_{x'}J_{dl}(x';\Theta)$
    \\
    9 &
    \qquad $dire = {\rm sign}(g_{t+1} + g'_{t+1})$
    \\
    10 &
    \qquad $p_{\emph{dire}} = {\rm SoftMax} (|g_{t+1} + g'_{t+1}|^{-1})$
    \\
    11 &
    \qquad \textbf{For} $a^{n\!s}\in A_{ns}$
    \\
    12 &
    \qquad \quad Generate a random number $p_{tmp}\in (0,1]$.
    \\
    13 &
    \qquad \quad \textbf{If} ($p_{tmp} < p_{\emph{dire}}[a^{n\!s}]$) \textbf{Then}
    \\
    14 &
    \qquad \qquad $x{[a^{n\!s}]}=x{[a^{n\!s}]} + dire{[a^{n\!s}]} \!\times\! step_{-}size_{l}$
    \\
    15 &
    \qquad \quad \textbf{End If}
    \\
    16 &
    \qquad \textbf{End For}
    \\
    17 &
    \qquad $x = {\rm Clip}(x, \mathbb{I}) $
    \\
    18 &
    \qquad {\small Create $<\!x,x'\!>$ s.t. $x{\rm [A_{s}]}\!\neq x'{\rm [A_{s}]}$, $x{\rm [A_{ns}]} \!= x'{\rm [A_{ns}]}$.}
    \\
    19 &
    \qquad \textbf{If} ($f(x;\Theta)\neq f(x';\Theta)$) \textbf{Then}
    \\
    20 &
    \qquad \quad $\Omega_{l}= \Omega_{l}\cup{<\!x,x'\!>}$
    \\
    21 &
    \qquad \textbf{End If}
    \\
    22 &
    \quad \textbf{End For}
    \\
    23 &
    \textbf{End For}
    \\
  \bottomrule
  \end{tabular}
  \vspace{-0.3cm}
\end{table}
}

\subsubsection{Global Generation}

To increases the IDIs' diversity,
we design a dynamic loss as follows:, % through combining different biased neurons,
\vspace{-0.2cm}
\begin{equation}\label{equ:dlCE}
\small
  J_{dl}(x;\Theta)  =
        - \frac{1}{N}
            \sum_{i=0}^{N-1}
                \sum_{k=1}^{N^{k}}
                    [ (p_{k}|r_{k}) \times f^{k}(x_{i}';\Theta) \times \log ( f^{k}(x_{i};\Theta) ) ]
\vspace{-0.2cm}
\end{equation}
where $x_{i}'$ comes from $x_{i}$ after flipping its sensitive attribute,
$N^{k}$ is the number of neurons in the most biased layer,
$f^{k}$ is the activation output of the $k$-th neuron.
$p$ is the position of biased neurons,
$r$ is the position of randomly selected neurons to increase the dynamics of $J_{dl}(x;\Theta)$.
$r=Rand_{(0,1)}(p_{r})$,
where $Rand_{(0,1)}(p_{r})$ returns a random vector with only `0' or `1'.
$r$ has the same size as $p$ and
satisfies $\sum{r}={\rm INT}(N^{k} \!\times\! p_{r})$,
where $\rm INT(\cdot)$ returns an integer.
Here, we set $p_{r}=5\%$.
`|' means `or',
$p_{k}|r_{k}\!=\!0$ if and only $p_{k}\!=\!0$ and $r_{k}\!=\!0$.
% $J_{dl}(\cdot;\cdot)$ gains dynamism by randomly adding several new neurons as the loss calculation.
The optimization object of IDI generation is:
\textbf{arg max} $J_{dl}(x;\Theta)$.

% is designed by DNN's hidden layer,
% and IDI determination of \emph{Definition 1} is based on DNN's output layer.
% The two criteria are inconsistent, thus the discrimination interpretation is more convincing by generating useful IDIs.
% %\cy{What does this paragraph mean?? Definition 1?}

\textbf{Algorithm 2} shows the details of global generation with momentum acceleration.
We first adopt k-means clustering function KMeans($X,N_{c}$) to process $X$ into $N_{c}$ clusters,
and then get seeds from clusters in a round-robin fashion at line 3.
We update random vector $r$ at equal intervals from lines 5 to 7,
not only to increase the dynamics but also to avoid excessively disturbing the generation task.
According to \emph{Definition 1},
we determine the IDIs from lines 8 to 12. % \cy{Definition 1? zhb:Yes}
We employ the momentum acceleration operation at lines 13 and 14,
which can effectively use historical gradient and reduce invalid searches.
Note that we keep the value of the sensitive attribute in $x$ at line 16.
Finally, we clip the value of $x$ to satisfy the input domain $\mathbb{I}$.

\subsubsection{Local Generation}
Since the local generation aims to find as many IDIs as possible near the seeds,
we increase the iteration number of each seed $max_{-}iter_{l}$,
and reduce the bias perturbation added in each iteration,
as shown in \textbf{Algorithm 3}.
Compared to the global phase, the major difference is the loop from lines 11 to 16,
where we add perturbation to the non-sensitive attributes of large gradients with a small probability.
We automatically get the probability of adding perturbation to each attribute in $x$ at line 10.

\subsection{Generalization Framework on Unstructured Data\label{ImageExtended}}

\begin{figure}[t]
\centering
    \includegraphics[width=\linewidth]{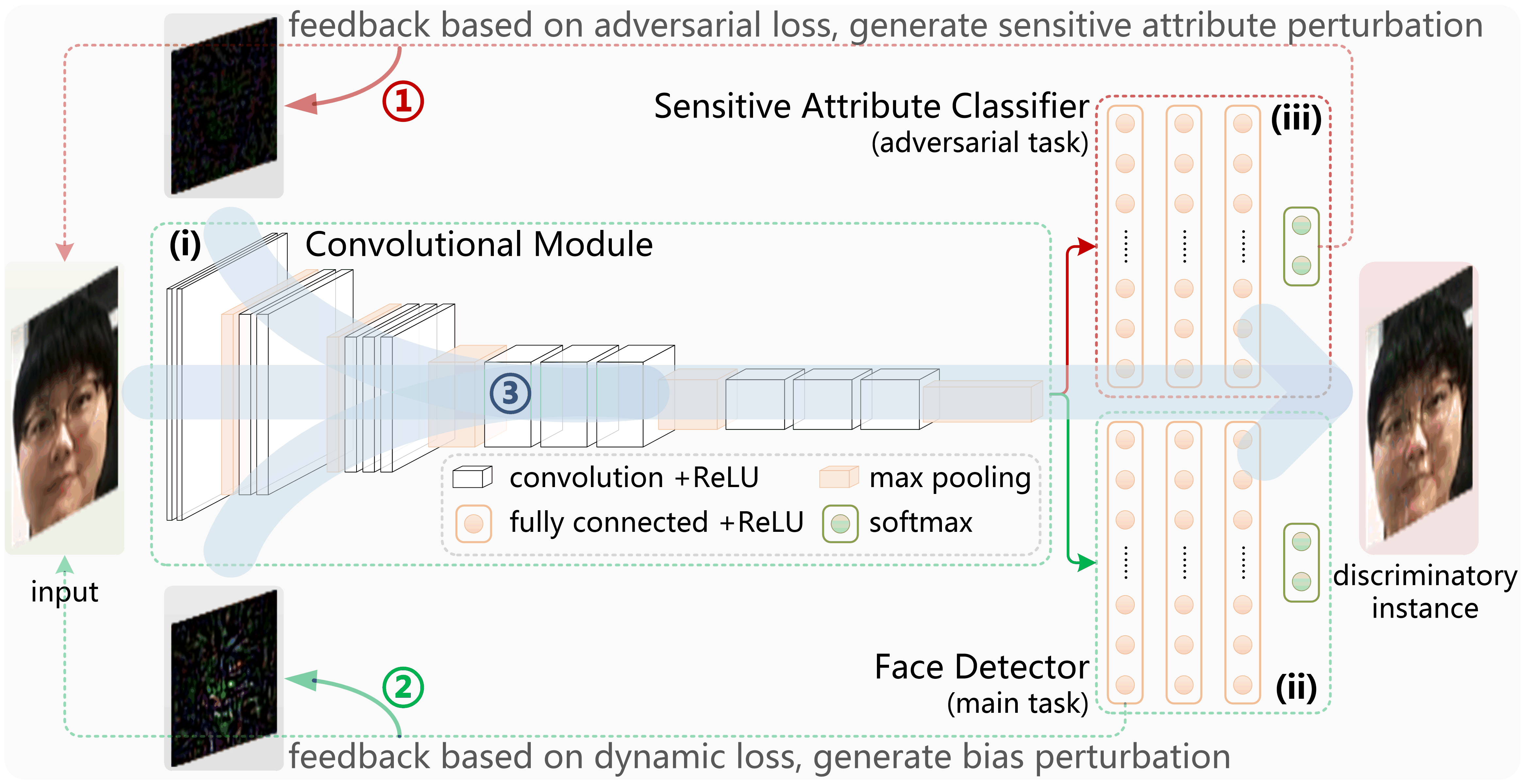}
  \caption{%An overview of fairness testing against a face image detector.
        An overview of generalization framework on image data type.
        }
  \label{fig:frameworkImageTest}
  \vspace{-0.4cm}
\end{figure}

% The sensitive attributes $A_{s}$ of unstructured data are not as clear as structured data,
We intend to solve the challenge of $A_{s}$ modification to generalize NeuronFair to unstructured data.
Here we take image data as an example.
Attributes of an image are determined by pixels with normalized values between 0 and 1,
i.e., the input domain of images is $\mathbb{I}\in [0,1]$.
Motivated by the adversarial attack,
we design a generalization framework to implement the image's $A_{s}$ modification,
which modifies $A_{s}$ through adding a small perturbation to most pixels,
as shown in Fig.~\ref{fig:frameworkImageTest}.

We consider a fairness testing scenario for face detection,
which determines whether the input image contains a face.
The face detector consists of a CNN module (i.e., Fig.~\ref{fig:frameworkImageTest} (i)) and a FCN module (i.e., Fig.~\ref{fig:frameworkImageTest} (ii)).
As shown in Fig.~\ref{fig:frameworkImageTest},
for a given face image $x$ and a detector $f(x,\Theta)$,
there are three steps:
{{\textcircled{\footnotesize 1}}} build a sensitive attribute classifier;
{{\textcircled{\footnotesize 2}}} produce $\Delta_{senatt}$ based on Eq.~(\ref{equ:FGSMsenatt}),
$\Delta_{senatt}$ is the perturbation added to image to flip sensitive attribute;
{{\textcircled{\footnotesize 3}}} generate $\Delta_{bias}$ based on NeuronFair,
where $\Delta_{bias}$ is the bias perturbation added to an image to flip the detection result.

First, we need a sensitive attribute classifier $f_{sa}(x;\Theta_{sa})$ that can distinguish the face image's $A_{s}$ (e.g., gender).
We build the $A_{s}$ classifier by adding a new FCN module (i.e., Fig.~\ref{fig:frameworkImageTest} (iii)) to the face detector's CNN module (i.e., Fig.~\ref{fig:frameworkImageTest} (i)).
Then, we froze the weights of the CNN module, and train the weights of the newly added FCN module.

Next, we modify the face image's $A_{s}$ based on the adversarial attack.
A classic adversarial attack FGSM~\cite{goodfellow2014explaining} is adopted to flip the predicted result of the sensitive attribute by generating $\Delta_{senatt}$ as follows:
\vspace{-0.1cm}
\begin{equation}\label{equ:FGSMsenatt}
    \begin{array}{c}
      \Delta_{senatt} = \epsilon \!\times\! {\rm sign}(\nabla_{x} J(x,y_{sa};\Theta_{sa})) \\
      {\rm satisfying~that}~~f_{sa}(x;\Theta_{sa}) \neq f_{sa}(x \!+\! \Delta_{senatt};\Theta_{sa})
    \end{array}
\vspace{-0.1cm}
\end{equation}
where $\epsilon$ is a hyper-parameter to determine perturbation size,
$\rm sign (\cdot)$ is a signum function return ``-1'', ``0'', or ``1'',
$x$ is an input image,
$y_{sa}$ is the ground-truth of $x$ about sensitive attributes,
$\Theta_{sa}$ is the weights of the $A_{s}$ classifier.

Finally, we leverage NeuronFair to generate $\Delta_{bias}$,
and then determine whether the instance pair $<\!x\!+\!\Delta_{bias},x\!+\!\Delta_{bias}\!+\!\Delta_{senatt}\!>$
satisfy \emph{Definition 1}.
We determine the discrimination at each layer of the detector at first.
For the CNN module,
the activation output of the convolutional layer is flattened.
Then, in the process of image IDI generation,
only the global generation is employed,
which is due to the different data forms between image and structured data.
% The dimension of the image is much higher than that of the structured data.
Taking the input image in Fig.~\ref{fig:frameworkImageTest} as an example,
its attributes can be regarded as $A\in \mathbb{R}^{{\rm 64 \times 64 \times 3}}$.
Based on a seed image IDI generated in the global phase,
numerous image IDIs will evolve in the local phase.
However, these image IDIs are similar,
with only a few pixel differences,
which have little effect on fairness improvement of the face detector.
Besides, cancel the signum function ${\rm sign}(\cdot)$ at line 15 of \textbf{Algorithm 2} for image data.

\section{Experimental Setting\label{Setup}}

\subsection{Datasets}

We evaluate NeuronFair on 7 datasets of which five are structured datasets and two are image datasets.
Each dataset is divided into three parts, i.e., 70\%, 10\%, 20\% as training, validation, and testing, respectively.

The 5 open-source structured datasets include
Adult,
German credit (GerCre),
bank marketing (BanMar),
COMPAS,
and medical expenditure panel survey (MEPS).
The details of these datasets are shown in Tab.~\ref{tab:textdata}.
All datasets can be downloaded from GitHub~\footnote{\url{https://github.com/Trusted-AI/AIF360/tree/master/aif360/data}}
and preprocessed by AI Fairness 360 toolkit (AIF360)~\cite{Bellamy2018AIF360}.

The 2 image datasets (i.e., ClbA-IN and LFW-IN) are constructed by ourselves for face detection.
ClbA-IN dataset consists of 60,000 face images from CelebA~\cite{Liu2015Deep} and 60,000 non-face images from ImageNet~\cite{Deng2009ImageNet}.
LFW-IN dataset consists of 10,000 face images from LFW~\cite{huang2008Labeled} and 10,000 non-face images from ImageNet~\cite{Deng2009ImageNet}.
The pixel value of each image is normalized to [0,1].

\subsection{Classifiers}
We implement 5 FCN-based classifiers~\cite{Zhang2020White,Bellamy2018AIF360} for structured datasets and
2 CNN-based face detectors~\cite{simonyan2014very,he2016deep} for image datasets
since FCN and CNN are the most widely used basic structures in real-world classification tasks.

The 5 FCN-based classifiers can be divided into two types.
The one is composed of 5 hidden layers for processing low-dimensional data (i.e., Adult, GerCre, BanMar),
denoted as LFCN.
The another is composed of 8 hidden layers for processing high-dimensional data (i.e., COMPAS and MEPS),
denoted as HFCN.
The activation functions in hidden layers and the output layer are ReLU and Softmax, respectively.
The hidden layer structures of LFCN and HFCN are
[64, 32, 16, 8, 4] and
[256, 256, 64, 64, 32, 32, 16, 8],
respectively.

The 2 CNN-based face detectors serve for face detection,
which are variants from two pre-trained models (i.e., VGG16~\cite{simonyan2014very} and ResNet50 \cite{he2016deep}) of \emph{keras.applications}.
We use the CNN module of VGG16 and ResNet50 as Fig.~\ref{fig:frameworkImageTest} (i),
and design the FCN module of Fig.~\ref{fig:frameworkImageTest} (ii) and (iii) as [512, 256, 128, 64, 16].

\subsection{Baselines}
We implement and compare 4 state-of-the-art (SOTA) methods with NeuronFair to evaluate their performance,
including
Aequitas~\cite{Udeshi2018Automated},
SymbGen~\cite{Aggarwal2018Automated},
ADF~\cite{Zhang2020White},
and EIDIG~\cite{Zhang2021Efficient}.
Note that
Themis~\cite{Galhotra2017Fairness} has been shown to be significantly less effective for DNN
and thus is omitted~\cite{Zhang2020White,Aggarwal2018Automated}.
We obtained the implementation of these baselines from GitHub~\footnote{\url{https://github.com/pxzhang94/ADF}}~\footnote{\url{https://github.com/LingfengZhang98/EIDIG}}.
All baselines are configured according to the best performance setting reported in the respective papers.

\begin{table}[t]
\centering
\caption{Details of the datasets.}
\vspace{-0.3cm}
\label{tab:textdata}
\resizebox{\linewidth}{!}{
{\Large
\begin{tabular}{lllrr}
\toprule
\hline
\textbf{Datasets} & \textbf{Scenarios} & \textbf{Sensitive Attributes} & \textbf{\# records} & \textbf{Dimensions} \\ \hline
Adult             & census income      & gender, race, age             & 48,842              & 13                  \\
GerCre            & credit             & gender, age                   & 1,000               & 20                  \\
BanMar            & credit             & age                           & 41,188              & 16                  \\
COMPAS            & law                & race                          & 5,278               & 400                 \\
MEPS              & medical care       & gender                        & 15,675              & 137                 \\ \hline
ClbA-IN              & face detection       & gender, race                        & 120,000              & 64$\times$64$\times$3                 \\
LFW-IN              & face detection       & gender, race                        & 20,000              & 64$\times$64$\times$3                 \\ \hline
\bottomrule
\end{tabular}
}
}
\vspace{-0.3cm}
\end{table}

\subsection{Evaluation Metrics}
Five aspects of NeuronFair are evaluated,
including
generation \textit{effectiveness},
\textit{efficiency},
\textit{interpretability},
the \textit{utility} of AUC metric, and
\textit{generalization} of NeuronFair.

\subsubsection{Generation Effectiveness Evaluation}
We evaluate the effectiveness of NeuronFair on structured data from two aspects:
generation quantity and quality.

(1) \textbf{Quantity}.
To evaluate the generation quantity,
we first count the total number of IDIs,
then count the global IDIs' number and local IDIs' number respectively,
recorded as `\#IDIs'.
Note that the duplicate instances are filtered.

(2) \textbf{Quality}.
We use
generation success rate (GSR),
generation diversity (GD), and
IDIs' contributions to fairness improvement (DM-RS~\cite{Udeshi2018Automated,Zhang2020White,Zhang2021Efficient})
to evaluate IDIs' quality.
% as follows:

\vspace{-0.2cm}
\begin{equation}\label{equ:GSR}
\centering
    \rm GSR=\frac{\#~IDIs}{\#~non\!-\!duplicate~instances}\times 100\%
\vspace{-0.1cm}
\end{equation}
where non-duplicate instances represent the input space.

\vspace{-0.2cm}
\begin{equation}\label{equ:GD}
\centering
    \rm GD_{NF}(\rho_{\rm cons},baseline) = \frac{CR_{NF-bl}}{CR_{bl-NF}}
\vspace{-0.1cm}
\end{equation}
where $\rm CR_{NF-bl}=\frac{\#~IDIs~of~baselines~fall~in~\Pi_{NF}}{\rm \#~IDIs~of~baseline}$ represents the coverage rate of the NeuronFair's IDIs to baseline's IDIs,
$\Pi_{\rm NF}$ is the area with NeuronFair's IDIs as the center and cosine distance $\rho_{\rm cons}$ as the radius;
similar to $\rm CR_{bl-NF} =\frac{\#~IDIs~of~NeuronFair~fall~in~\Pi_{bl}}{\rm \#~IDIs~of~NeuronFair}$.
The NeuronFair's IDIs are more diverse
when ${\rm GD_{NF}}>$1.

The generated IDIs serve to improve DNN's fairness by using these IDIs to retrain it.
DM-RS is the percentage of IDIs in randomly sampled instances.
High DM-RS value represents that the DNN is biased,
i.e., the IDI's contribution to fairness improvement is low.

\vspace{-0.1cm}
\begin{equation}\label{equ:DM-RS}
\centering
    \rm DM\!-\!RS = \frac{\#~IDIs}{\#~instances~randomly~sampled} \times 100\%
\vspace{-0.1cm}
\end{equation}

\subsubsection{Efficiency Evaluation}
We evaluate the efficiency of NeuronFair by generation speed~\cite{Zhang2020White},
i.e., the time cost of generating 1,000 IDIs (\#sec/1,000 IDIs).

\subsubsection{Interpretability Evaluation based on Biased Neurons}
To interpret the utility of NeuronFair,
we refer to paper~\cite{Pei2019DeepXplore} to design the coverage of biased neurons,
which is defined as follows:
for a given instance,
compute the activation output of the most biased layer;
normalize the activation values;
select neurons with activation values greater than 0.5 as the activated neurons;
compare the coverage of the activated neurons to the biased neurons.

\subsubsection{Utility Evaluation of AUC Metric}

In our work, based on the interpretation results,
we design AUC value to measure the discrimination.
We evaluate the utility of AUC metrics from three aspects:
consistency, significance, and complexity between AUC and DM-RS.

(1) \textbf{Consistency}.
To evaluate the consistency, we adopt Spearman's correlation coefficient, as follows:
\vspace{-0.1cm}
\begin{equation}\label{equ:SPCC}
\centering
    \rho_{s} = 1 - \frac{6\sum_{i=1}^{n}d_{i}^{2}}{n(n^{2}-1)}
\vspace{-0.1cm}
\end{equation}
where $d_{i} = a_{i}-b_{i}$,
$i\in\{1,2,...,n\}$,
$a_{i}$ and $b_{i}$ are the rank of AUC and DM-RS values, respectively.
High $\rho_{s}$ means more consistent.

(2) \textbf{Significance}.
To evaluate whether AUC can measure discrimination more significantly than DM-RS,
we use the standard deviation, as follows:
\vspace{-0.1cm}
\begin{equation}\label{equ:StanD}
\centering
    \sigma = \sqrt{\frac{\sum_{i=1}{n}(c_{i}-\mu)^{2}}{n-1}}
\vspace{-0.1cm}
\end{equation}
where $c_{i}$ is AUC or DM-RS of different testing methods,
$i\in\{1,2,...,n\}$,
$\mu$ is the mean value of $c_{i}$.
Large $\sigma$ means more significant.

\subsubsection{Generalization Evaluation on Image Data}
We evaluate the generalization of NeuronFair on image data from two aspects:
generation quantity, and quality.

(1) \textbf{Quantity}.
To evaluate the generation quantity on image data,
we only count the global image IDIs' number,
recorded as `\#IDIs'.

(2) \textbf{Quality}.
We adopt
GSR and
IDIs' contributions to face detector's fairness improvement based on AUC value
to evaluate IDIs' quality,
then compute its detection rate (DR) after retraining.

\subsection{Implementation Details}

To fairly study the performance of the baselines and NeuronFair,
our experiments have the following settings:
(1)~the hyperparameters of each method are set according to Tab.~\ref{tab:Parameters},
where `Glo.' and `Loc.' represent the global and local phases, respectively;
(2)~for the FCN-based classifier,
we set the learning rate to 0.001,
and choose Adam as the optimizer;
for the CNN-based face detector,
we set the learning rate to 0.01,
and choose SGD as the optimizer;
the training results are shown in Tab.~\ref{tab:modelTraining},
where {``99.83\%/92.80\%/94.30\%''} represents the accuracy of face detector, gender classifier, and race classifier, respectively.

We conduct all the experiments on a server with
one Intel i7-7700K CPU running at 4.20GHz,
64 GB DDR4 memory,
4 TB HDD and
one TITAN Xp 12 GB GPU card.

\begin{table}[t]
\caption{Parameter setting of experiments.
        }
\vspace{-0.3cm}
\label{tab:Parameters}
\centering
\resizebox{\linewidth}{!}{
{\huge
\begin{tabular}{clrcll}
\toprule[0.5mm]
\hline
\textbf{No.} & \multicolumn{1}{c}{\textbf{Parameters~~}} & \multicolumn{3}{c}{\textbf{Values (Glo.~/~Loc.)}} & \multicolumn{1}{c}{\textbf{Descriptions}}      \\ \hline
1            & $N_{c}$                                 & 4               & /           &  {\Large \XSolidBrush} & the number of clusters for global generation            \\
2            & $num_{g}$                               & 1,000           & /           &  {\Large \XSolidBrush} & the number of seeds for global generation      \\
3            & $max_{-}iter$                           & 40              & /           & 1,000           & the maximum number of iterations for each seed \\
4            & $step_{-}size$                          & 1.0             & /           & 1.0             & the perturbation size of each iteration        \\
5            & $\mu$                                   & 0.1             & /           & 0.05            & the decay rate of momentum, $\mu\in(0.01,0.20)$                     \\
6            & $r_{-}step$                             & 10              & /           & 50              & the step size for random disturbance, $r\!_{-}step\in\!(5,100)$          \\ \hline
\bottomrule[0.5mm]
\end{tabular}
}
}
\vspace{-0.3cm}
\end{table}

\begin{table}[t]
\centering
\caption{The accuracy of classifiers and face detectors.
        % where {\scriptsize{``99.83\%/94.30\%''}} represents the accuracy of face detector and race classifier, respectively.
        }
\vspace{-0.3cm}
\label{tab:modelTraining}
\resizebox{\linewidth}{!}{
{\Large
\begin{tabular}{lccccccc}
\toprule
\hline
\textbf{Datasets} & Adult   & GerCre   & BanMar  & COMPAS  & MEPS    & ClbA-IN                                                               & LFW-IN                                                                \\ \hline
\textbf{Classifiers}     & LFC-A   & LFC-G    & LFC-B   & HFC-C   & HFC-M   & VGG16                                                                 & ResNet50                                                              \\ \hline
\textbf{accuracy} & 88.36\% & 100.00\% & 96.71\% & 92.20\% & 98.13\% & \begin{tabular}[c]{@{}c@{}}99.83\%/\\ 92.80\%/\\ 94.30\%\end{tabular} & \begin{tabular}[c]{@{}c@{}}99.56\%/\\ 94.40\%/\\ 94.20\%\end{tabular} \\
\hline
\bottomrule
\end{tabular}
}
}
\vspace{-0.3cm}
\end{table}

\section{Experimental Results\label{Experiments}}
We evaluate NeuronFair through answering the following five research questions (RQ):
(1) how \textit{effective} is NeuronFair;
(2) how \textit{efficient} is NeuronFair;
(3) how to \textit{interpret} the utility of NeuronFair;
(4) how \textit{useful} is the AUC metric;
(5) how \textit{generic} is NeuronFair?

\subsection{Research Questions 1}
%%%%%%%%%%%%%%%%%%%%%%%%%%%%%%%%%%%%%%%% RQ1 %%%%%%%%%%%%%%%%%%%%%%%%%%%%%%%%%%%%%%%%%
% \vspace{0.3cm}
\vspace{-0.3cm}
\begin{center}
\fcolorbox{black}{gray!20}{\parbox{0.97\linewidth}
    {
        % \emph{\textbf{RQ2}}: How effective and efficient is NeuronFair in generating IDIs?
        How effective is NeuronFair in generating IDIs?
    }
}
\end{center}

When reporting the results,
we focus on the following aspects:
generation \textit{quantity} and \textit{quality}.

%%%%%%%%%%%%%%%%%%%%%%%%%%%%%%%%%%%%%%%%%比较数量
\textbf{Generation Quantity}.
The evaluation results are shown in Tabs.~\ref{tab:comparisonAequitasADF},
\ref{tab:comparisonSymbGen}, and
\ref{tab:IDIGlobalLoacl},
including three scenarios:
the \textit{total} number of IDIs,
the IDIs number in \textit{global} phase,
and the IDIs number in \textit{local} phase.
% From the results, we have the following observations.

Implementation details for quantity evaluation:
(1)~SymbGen works differently from other baselines,
thus we follow the comparison strategy of Zhang et al.~\cite{Zhang2020White},
i.e., evaluating the generation quantity of NeuronFair and SymbGen within the same time (i.e., 500 sec) limit, as shown in Tab.~\ref{tab:comparisonSymbGen};
(2)~for a fair global phase comparison,
we generate 1,000 non-duplicate instances without constrained by $num_{g}$,
then count IDIs number and record it in Tab.~\ref{tab:IDIGlobalLoacl},
where the seeds used are consistent for different methods;
(3)~for a fair local phase comparison,
we mix IDIs generated globally by different methods,
and randomly sample 100 as the seeds in local phase;
then generate 1,000 non-duplicate instances for each seed without constrained by $max_{-}iter_{l}$,
count the IDIs number on average for each seed and record it in Tab.~\ref{tab:IDIGlobalLoacl}.

\begin{itemize}[leftmargin=9pt,topsep=0pt,partopsep=0pt]
    \item % 比较所有的数量
NeuronFair generates more IDIs than baselines,
especially for densely coded structured data.
For instance,
in Tab.~\ref{tab:comparisonAequitasADF},
on Adult dataset with different attributes,
the IDIs number of NeuronFair is 217,855 on average,
which is 16.5 times and 1.6 times that of Aequitas and EIDIG, respectively.
In addition, in Tab.~\ref{tab:comparisonSymbGen},
NeuronFair generates much more IDIs than SymbGen on all datasets.
The outstanding performance of NeuronFair
is mainly because the optimization object of NeuronFair
takes into account the whole DNNs' discrimination information through the biased neurons
while Aequitas and EIDIG only depend on the output layer.
However, the IDIs number on COMPAS dataset with race gender
is 11,232,
which is slightly lower than that of EIDIG.
Since the COMPAS is encoded as one-hot in AIF360~\cite{Bellamy2018AIF360},
we speculate the reason is that
too sparse data coding reduces the derivation efficiency from biased neurons.
\end{itemize}

\begin{table}[t]
\caption{Comparison with Aequitas, ADF, and EIDIG based on the total number of generated IDIs.}
\vspace{-0.3cm}
\label{tab:comparisonAequitasADF}
\resizebox{\linewidth}{!}{ \huge
\begin{tabular}{llrrrrrrrr}
\toprule[0.5mm]
\hline
\multirow{2}{*}{\textbf{Datasets}} & \multirow{2}{*}{\textbf{\begin{tabular}[c]{@{}l@{}}Sen.\\ Att.\end{tabular}}} & \multicolumn{2}{c}{\textbf{Aequitas}}                   & \multicolumn{2}{c}{\textbf{ADF}}                    & \multicolumn{2}{c}{\textbf{EIDIG}}                  & \multicolumn{2}{c}{\textbf{NeuronFair}} \\ \cline{3-10}
                                   &                                                                               & \textbf{\#IDIs} & \multicolumn{1}{r|}{\textbf{GSR}}     & \textbf{\#IDIs} & \multicolumn{1}{r|}{\textbf{GSR}} & \textbf{\#IDIs} & \multicolumn{1}{r|}{\textbf{GSR}} & \textbf{\#IDIs}    & \textbf{GSR}       \\ \hline
\multirow{3}{*}{Adult}             & gender                                                                        & 1,995           & \multicolumn{1}{r|}{8.35\%}           & 33,365          & \multicolumn{1}{r|}{16.42\%}      & 57,386          & \multicolumn{1}{r|}{27.24\%}      & \textbf{122,370}   & \textbf{28.19\%}   \\
                                   & race                                                                          & 13,132          & \multicolumn{1}{r|}{8.65\%}           & 57,716          & \multicolumn{1}{r|}{23.32\%}      & 88,650          & \multicolumn{1}{r|}{32.81\%}      & \textbf{172,995}   & \textbf{34.19\%}   \\
                                   & age                                                                           & 24,495          & \multicolumn{1}{r|}{10.48\%}          & 188,057         & \multicolumn{1}{r|}{46.94\%}      & 251,156         & \multicolumn{1}{r|}{48.69\%}      & \textbf{358,201}   & \textbf{49.39\%}   \\ \cline{2-10}
\multirow{2}{*}{GerCre}            & gender                                                                        & 4,347           & \multicolumn{1}{r|}{15.24\%}          & 57,386          & \multicolumn{1}{r|}{15.43\%}      & 64,075          & \multicolumn{1}{r|}{17.23\%}      & \textbf{68,218}    & \textbf{36.57\%}   \\
                                   & age                                                                           & 44,800          & \multicolumn{1}{r|}{38.63\%}          & 236,551         & \multicolumn{1}{r|}{58.74\%}      & 239,107         & \multicolumn{1}{r|}{59.38\%}      & \textbf{255,971}   & \textbf{63.35\%}   \\ \cline{2-10}
BanMar                             & age                                                                           & 10,138          & \multicolumn{1}{r|}{27.21\%}          & 167,361         & \multicolumn{1}{r|}{30.75\%}      & 197,341         & \multicolumn{1}{r|}{36.26\%}      & \textbf{302,821}   & \textbf{47.76\%}   \\ \cline{2-10}
COMPAS                             & race                                                                          & 658             & \multicolumn{1}{r|}{\textbf{18.87\%}} & 12,335          & \multicolumn{1}{r|}{2.22\%}       & \textbf{13,451} & \multicolumn{1}{r|}{2.32\%}       & 11,232             & 1.62\%             \\ \cline{2-10}
MEPS                               & gender                                                                        & 6,132           & \multicolumn{1}{r|}{13.51\%}          & 77,794          & \multicolumn{1}{r|}{16.37\%}      & 101,132         & \multicolumn{1}{r|}{21.28\%}      & \textbf{130,898}   & \textbf{27.91\%}   \\ \hline \bottomrule[0.5mm]
\end{tabular}
}
\vspace{-0.3cm}
\end{table}

\begin{table}[t]
\caption{Comparison with SymbGen based on the number of IDIs generated in 500 seconds.}
\vspace{-0.3cm}
\label{tab:comparisonSymbGen}
{% \tiny
\resizebox{0.75\linewidth}{!} {
\begin{tabular}{llrrrr}
\toprule
\hline
\multirow{2}{*}{\textbf{Datasets}} & \multirow{2}{*}{\textbf{Sen. Att.}} & \multicolumn{2}{c}{\textbf{\qquad SymbGen\qquad}}                                            & \multicolumn{2}{c}{\textbf{\qquad NeuronFair\qquad}}                                   \\ \cline{3-6}
                                   &                                     & \multicolumn{1}{r}{\quad \textbf{\#IDIs}} & \multicolumn{1}{r|}{\quad\textbf{GSR}} & \multicolumn{1}{r}{\quad\textbf{\#IDIs}} & \multicolumn{1}{r}{\quad\textbf{GSR}} \\ \hline
\multirow{3}{*}{Adult}             & gender                              & 195                                 & \multicolumn{1}{r|}{13.89\%}         & \textbf{4,048}                      & \textbf{25.24\%}                    \\
                                   & race                                & 452                                 & \multicolumn{1}{r|}{11.01\%}         & \textbf{4,532}                      & \textbf{39.54\%}                    \\
                                   & age                                 & 531                                 & \multicolumn{1}{r|}{12.17\%}         & \textbf{5,760}                      & \textbf{50.74\%}                    \\ \cline{2-6}
\multirow{2}{*}{GerCre}            & gender                              & 821                                 & \multicolumn{1}{r|}{18.92\%}         & \textbf{3,610}                      & \textbf{27.55\%}                    \\
                                   & age                                 & 1,034                               & \multicolumn{1}{r|}{37.19\%}         & \textbf{3,796}                      & \textbf{51.40\%}                    \\ \cline{2-6}
BanMar                             & age                                 & 672                                 & \multicolumn{1}{r|}{30.79\%}         & \textbf{3,095}                      & \textbf{56.79\%}                    \\ \cline{2-6}
COMPAS                             & race                                & 42                                  & \multicolumn{1}{r|}{1.33\%}          & \textbf{124}                        & \textbf{2.08\%}                     \\ \cline{2-6}
MEPS                               & gender                              & 404                                 & \multicolumn{1}{r|}{14.22\%}         & \textbf{3,252}                      & \textbf{26.35\%}                    \\ \hline
\bottomrule
\end{tabular}
}
}
\vspace{-0.3cm}
\end{table}

\begin{table}[t]
\caption{`\#IDIs' measurement in the global and local phases.}
\vspace{-0.3cm}
\label{tab:IDIGlobalLoacl}
\resizebox{\linewidth}{!}{ \Huge
\begin{tabular}{llrrrrrrrrrr}
\toprule[0.5mm]
\hline
\multirow{2}{*}{\textbf{Datasets}} & \multirow{2}{*}{\textbf{\begin{tabular}[c]{@{}l@{}}Sen.\\ Att.\end{tabular}}} & \multicolumn{5}{c}{\textbf{Global Phase}}                                                                                                                                                                                                                                                                                  & \multicolumn{5}{c}{\textbf{Local Phase}}                                                                                                                                                                                                                                                                                   \\ \cline{3-12}
                                   &                                                                               & \multicolumn{1}{c}{\textbf{\begin{tabular}[c]{@{}c@{}}Aequi\\ tas\end{tabular}}} & \multicolumn{1}{c}{\textbf{\begin{tabular}[c]{@{}c@{}}Symb\\ Gen\end{tabular}}} & \multicolumn{1}{c}{\textbf{ADF}} & \multicolumn{1}{c}{\textbf{EIDIG}} & \multicolumn{1}{c|}{\textbf{\begin{tabular}[c]{@{}c@{}}Neuron\\ Fair\end{tabular}}} & \multicolumn{1}{c}{\textbf{\begin{tabular}[c]{@{}c@{}}Aequi\\ tas\end{tabular}}} & \multicolumn{1}{c}{\textbf{\begin{tabular}[c]{@{}c@{}}Symb\\ Gen\end{tabular}}} & \multicolumn{1}{c}{\textbf{ADF}} & \multicolumn{1}{c}{\textbf{EIDIG}} & \multicolumn{1}{c}{\textbf{\begin{tabular}[c]{@{}c@{}}Neuron\\ Fair\end{tabular}}} \\ \hline
\multirow{3}{*}{Adult}             & gender                                                                        & 35                                                                               & 51                                                                              & 261                              & 404                                & \multicolumn{1}{r|}{\textbf{864}}                                                   & 57                                                                               & 63                                                                              & 128                              & 142                                & \textbf{143}                                                                       \\
                                   & race                                                                          & 98                                                                               & 143                                                                             & 332                              & 459                                & \multicolumn{1}{r|}{\textbf{959}}                                                   & 134                                                                              & 158                                                                             & 174                              & \textbf{193}                                & 189                                                                       \\
                                   & age                                                                           & 115                                                                              & 331                                                                             & 538                              & 695                                & \multicolumn{1}{r|}{\textbf{974}}                                                   & 213                                                                              & 267                                                                             & 350                              & 361                                & \textbf{367}                                                                       \\ \cline{2-12}
\multirow{2}{*}{GerCre}            & gender                                                                        & 69                                                                               & 128                                                                             & 541                              & 577                                & \multicolumn{1}{r|}{\textbf{599}}                                                   & 63                                                                               & 86                                                                              & 106                              & 111                                & \textbf{113}                                                                       \\
                                   & age                                                                           & 175                                                                              & 247                                                                             & 598                              & 599                                & \multicolumn{1}{r|}{\textbf{600}}                                                   & 256                                                                              & 301                                                                             & 396                              & 400                                & \textbf{426}                                                                       \\ \cline{2-12}
BanMar                             & age                                                                           & 74                                                                               & 244                                                                             & 678                              & 697                                & \multicolumn{1}{r|}{\textbf{999}}                                                   & 137                                                                              & 198                                                                             & 247                              & 283                                & \textbf{303}                                                                       \\ \cline{2-12}
COMPAS                             & race                                                                          & 94                                                                               & 187                                                                             & 745                              & 749                                & \multicolumn{1}{r|}{\textbf{930}}                                                   & 7                                                                                & 6                                                                               & 17                      & \textbf{18}                                 & 12                                                                                 \\ \cline{2-12}
MEPS                               & gender                                                                        & 73                                                                               & 210                                                                             & 650                              & 692                                & \multicolumn{1}{r|}{\textbf{1,000}}                                                 & 84                                                                               & 92                                                                              & 120                              & 146                                & \textbf{149}                                                                       \\ \hline \bottomrule[0.5mm]
\end{tabular}
}
\vspace{-0.3cm}
\end{table}

\begin{itemize}[leftmargin=9pt,topsep=0pt,partopsep=0pt]
    \item  % 比较全局的数量
As shown in Tab.~\ref{tab:IDIGlobalLoacl},
NeuronFair generates much more IDIs than all baselines in the global phase,
which is beneficial to increase the diversity of NeuronFair's IDIs in the subsequent local phase.
For instance,
on all datasets,
the IDIs number of NeuronFair is 866 on average,
which is 9.45 times and 1.42 times that of Aequitas and EIDIG, respectively.
This is mainly because the optimization object of NeuronFair
takes into account the dynamics through the dynamic combination of biased neurons.
Thus, NeuronFair searches a larger space to generate more global IDIs.
\end{itemize}

\begin{itemize}[leftmargin=9pt,topsep=0pt,partopsep=0pt]
    \item  % 比较局部的数量
In the local phase,
NeuronFair is much more efficient than baselines in general.
For instance,
in Tab.~\ref{tab:IDIGlobalLoacl},
on average,
NeuronFair returns 78.97\%, 45.35\%, 10.81\%, and 2.90\% more IDIs than Aequitas, SymbGen, ADF, and EIDIG, respectively.
Recall that Aequitas, ADF, EIDIG, and NeuronFair all guide local phase through a probability distribution,
which is the likelihood of IDIs by modifying several certain attributes
(i.e., the loop from lines 11 to 16 of \textbf{Algorithm 3}).
The probability determination of NeuronFair
takes into account the momentum and SoftMax activation (i.e., at line 10 of \textbf{Algorithm 3}) while
the baselines do not.
% while Aequitas applies the same probability to all instances,
% ADF and EIDIG determine the probability based on gradient.
Hence, NeuronFair generates more local IDIs.
% Besides, the local IDIs number of NeuronFair on the COMPAS dataset is not the best.
% Since the COMPAS is
% we speculate that it is caused by data sparsity.
\end{itemize}

%%%%%%%%%%%%%%%%%%%%%%%%%%%%%%%%%%%%%%%%%%比较质量

\begin{figure}[t]
\centering
\subfigure[Comparison with Aequitas]{
    \includegraphics[width=0.46\linewidth]{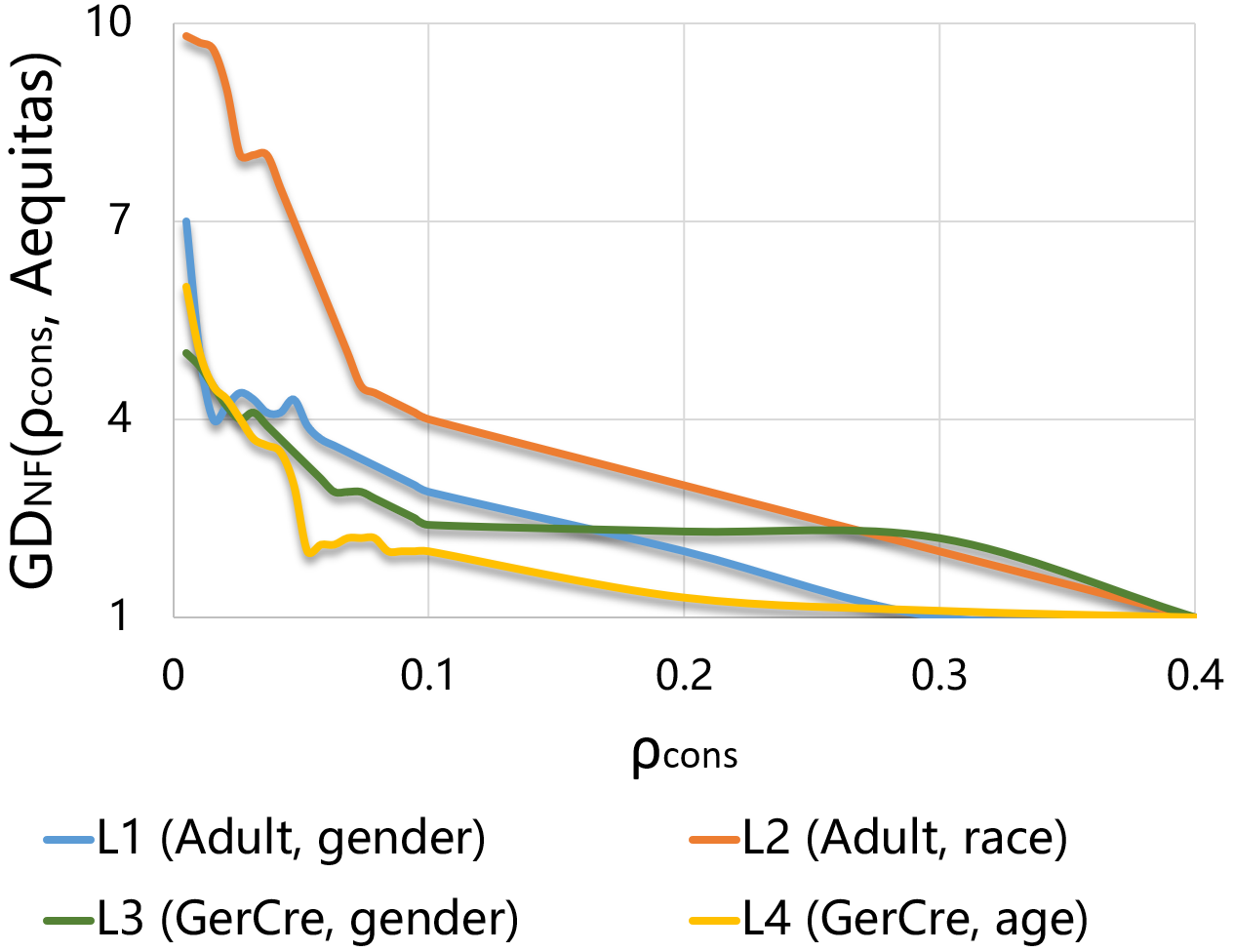}
    }
\subfigure[Comparison with ADF]{
    \includegraphics[width=0.46\linewidth]{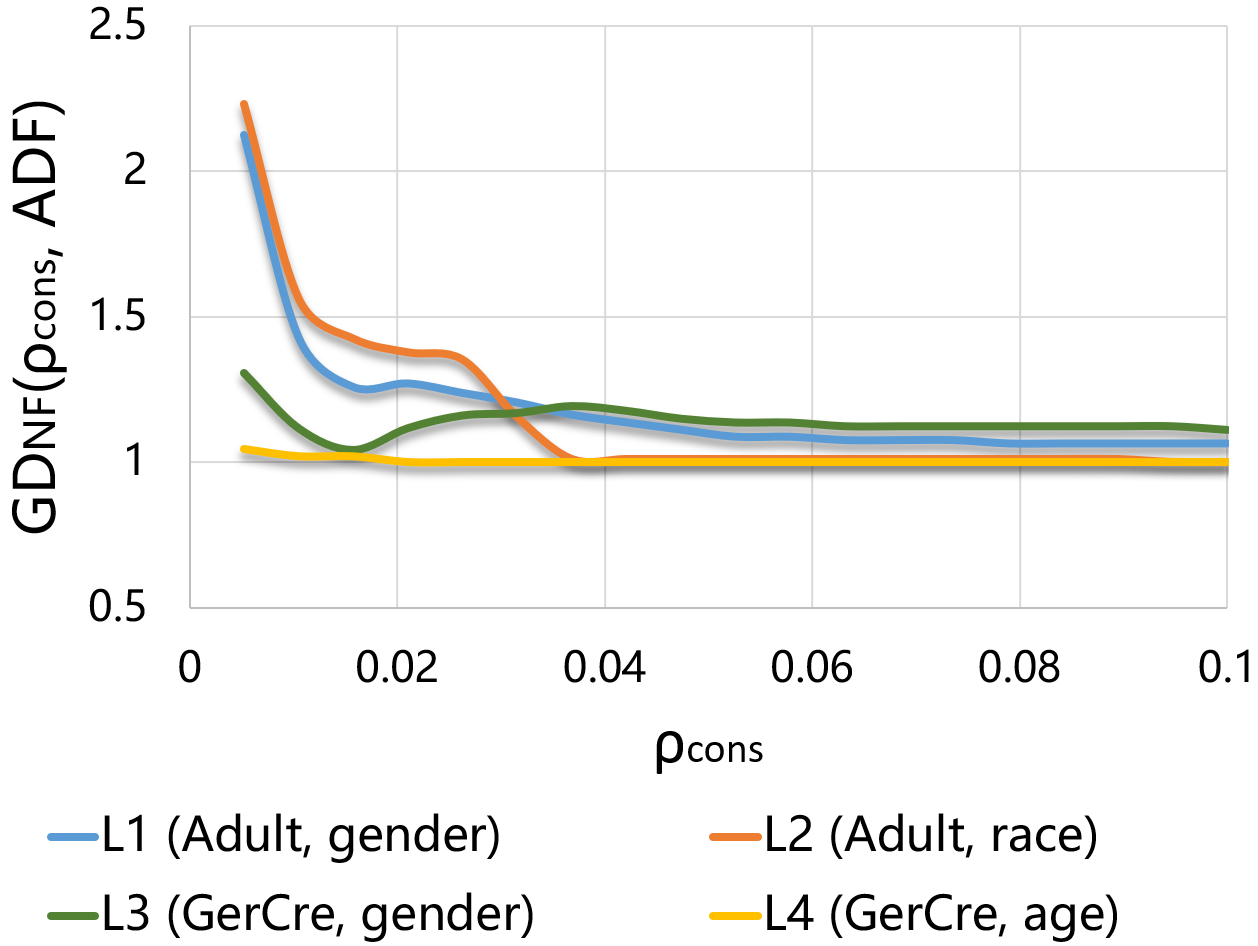}
    }
    \vspace{-0.4cm}
  \caption{Generation diversity of NeuronFair compared to Aequitas (left) and ADF (right).
        }
  \label{fig:coverRate}
  \vspace{-0.3cm}
\end{figure}

\textbf{Generation Quality}.
The evaluation results are shown in
Tabs.~\ref{tab:comparisonAequitasADF},
\ref{tab:comparisonSymbGen},
\ref{tab:FairnessImprovement}, and
Fig.~\ref{fig:coverRate},
including
the generation success rate (\textit{GSR}),
generation diversity (\textit{GD}),
and fairness improvement (\textit{DM-RS}).
% From the results, we have the following observations.

Implementation details for quality evaluation:
(1)~for a fair diversity comparison,
we seed each method with the same set of 10 global IDIs and apply them to generate 100 local IDIs for each seed without considering $max_{-}iter_{l}$,
as shown in Fig.~\ref{fig:coverRate};
(2)~we randomly select 10\% IDIs of each method to retrain DNNs,
then compute their fairness improvement results;
to avoid contingency,
we repeat 5 times and record the average DM-RS value in Tab.~\ref{tab:FairnessImprovement}.

\begin{itemize}[leftmargin=9pt,topsep=0pt,partopsep=0pt]
    \item  % 比较生成成功率
As shown in Tabs.~\ref{tab:comparisonAequitasADF} and \ref{tab:comparisonSymbGen},
the GSR values of NeuronFair are higher than that of baselines on almost all datasets,
i.e., NeuronFair can search for a larger valid input space,
where the input space is calculated by `\#IDIs/GSR'.
For instance,
in Tab.~\ref{tab:comparisonAequitasADF},
on all datasets,
Aequitas has a GSR of 17.62\% on average,
whereas NeuronFair achieves a GSR of 36.12\%,
which is $\sim\!\!\times$2.1 more than that of Aequitas.
The outstanding performance of NeuronFair
is mainly because it
not only considers the whole DNN's discrimination through biased neurons,
but also takes into account the dynamics of the optimization object through the combination of biased neurons.
Thus, NeuronFair searches a larger valid input space than Aequitas.

\quad Meanwhile,
the GSR value of NeuronFair on different sensitive attributes is more robust than ADF and EIDIG.
For instance,
in Tab.~\ref{tab:comparisonAequitasADF},
on the GerCre dataset with gender and age attributes,
the GSR values of NeuronFair are 36.57\% and 63.35\%,
whereas that of EIDIG are 17.23\% and 59.38\%.
We speculate the reason is that the discrimination about the gender attribute in the output layer is not obvious,
but NeuronFair can find potential fairness violations through the internal discrimination information of biased neurons.
Therefore, we can realize stable testing for different sensitive attributes.

\quad In addition, the valid input space of NeuronFair is larger than baselines in general,
i.e.,
a larger input space supports more diverse IDI generation.
For instance,
in Tab.~\ref{tab:comparisonSymbGen},
the average input space of NeuronFair is 3.30 times that of SymbGen.
It is mainly because the momentum acceleration strategy employs historical gradient as auxiliary guidance,
which reduces the number of invalid searches.
Hence, NeuronFair generates more IDIs in a large input space.
\end{itemize}

\begin{table}[t]
\caption{
        Fairness improvement measured by DM-RS,
        where `Before' and `After' represent the original and the retrained DNNs, respectively.
        }
\vspace{-0.3cm}
\label{tab:FairnessImprovement}
\resizebox{\linewidth}{!}{
\begin{tabular}{llrrrrrr}
\toprule
\hline
\multirow{2}{*}{\textbf{Datasets}} & \multirow{2}{*}{\textbf{\begin{tabular}[c]{@{}l@{}}Sen. \\ Att.\end{tabular}}} & \multirow{2}{*}{\textbf{Before}} & \multicolumn{5}{c}{\textbf{After}}                                                                                                                                                                                                                                                          \\ \cline{4-8}
                                   &                                                                                &                                  & \multicolumn{1}{c}{\begin{tabular}[c]{@{}c@{}}Aequi\\ tas\end{tabular}} & \multicolumn{1}{c}{\begin{tabular}[c]{@{}c@{}}Symb\\ Gen\end{tabular}} & \multicolumn{1}{c}{ADF} & \multicolumn{1}{c}{EIDIG} & \multicolumn{1}{c}{\textbf{\begin{tabular}[c]{@{}c@{}}Neuron\\ Fair\end{tabular}}} \\ \hline
\multirow{3}{*}{Adult}             & gender                                                                         & 2.88\%                           & 0.45\%                                                                  & 0.44\%                                                                 & 0.26\%                  & 0.21\%                    & \textbf{0.19\%}                                                                    \\
                                   & race                                                                           & 8.91\%                           & 0.61\%                                                                  & 0.81\%                                                                 & 0.75\%                  & 0.69\%                    & \textbf{0.57\%}                                                                    \\
                                   & age                                                                            & 14.56\%                          & 4.40\%                                                                  & 4.38\%                                                                 & 4.18\%                  & 3.74\%                    & \textbf{3.30\%}                                                                    \\ \cline{2-8}
\multirow{2}{*}{GerCre}            & gender                                                                         & 5.16\%                           & 0.76\%                                                                  & 0.67\%                                                                 & 0.55\%                  & 0.56\%                    & \textbf{0.49\%}                                                                    \\
                                   & age                                                                            & 30.90\%                          & 3.66\%                                                                  & 3.46\%                                                                 & 3.31\%                  & 3.21\%                    & \textbf{2.32\%}                                                                    \\ \cline{2-8}
BanMar                             & age                                                                            & 1.38\%                           & 0.68\%                                                                  & 0.52\%                                                                 & 0.76\%                  & 0.55\%                    & \textbf{0.39\%}                                                                    \\ \cline{2-8}
COMPAS                             & race                                                                           & 2.03\%                           & 1.48\%                                                                  & 1.20\%                                                                 & 0.75\%                  & 0.76\%                    & \textbf{0.52\%}                                                                    \\ \cline{2-8}
MEPS                               & gender                                                                         & 5.10\%                           & 1.30\%                                                                  & 2.15\%                                                                 & 1.28\%                  & 1.27\%                    & \textbf{1.26\%}                                                                    \\ \hline \bottomrule
\end{tabular}
}
\vspace{-0.3cm}
\end{table}

\begin{itemize}[leftmargin=9pt,topsep=0pt,partopsep=0pt]
    \item  % 比较多样性
In all cases,
NeuronFair can generate more diverse IDIs,
which is beneficial to discover more potential discrimination and then improve fairness through retraining.
For instance,
in Fig.~\ref{fig:coverRate},
compare to Aequitas and ADF,
the $\rm GD_{NF}$ values are all greater than `1' under different radius values $\rho_{\rm cons}$,
and as the radius increases, the value of $\rm GD_{NF}$ gradually converges to `1'.
It demonstrates that the IDIs generated by NeuronFair can always cover that of baselines.
We speculate the reason is that the dynamic loss function expands the valid input space by combining different biased neurons as the optimization object.

\quad Besides,
a close investigation shows that there is a similar trend in the generation diversity for the same sensitive attribute.
For instance,
% in Fig.~\ref{fig:coverRate},
when $\rho_{\rm cons}<$0.1 in Fig.~\ref{fig:coverRate} (a) or
$\rho_{\rm cons}<$0.02 in Fig.~\ref{fig:coverRate} (b),
the line `L2' with race is always the highest,
the line `L4' with age is always the lowest,
while lines `L1' and `L3' with gender are in the middle.
Since both datasets Adult and GerCre are related to money (i.e., salary and loans),
we speculate that there is similar discrimination for gender in classifiers LFC-A and LFC-G for similar tasks,
thus $\rm GD_{NF}$ shows similar trends in gender attribute.
\end{itemize}

\begin{itemize}[leftmargin=9pt,topsep=0pt,partopsep=0pt]
    \item  % 比较公平提升效果
In all cases, NeuronFair can obtain larger DM-RS values,
i.e.,
the IDIs generated by NeuronFair contribute more to the DNNs' fairness improvement.
For instance,
in Tab.~\ref{tab:FairnessImprovement},
measured by DM-RS,
NeuronFair realizes fairness improvement of 87.24\% on average,
versus baselines, i.e., 81.18\%	for Aequitas, 80.78\% for SymbGen, 83.30\% for ADF, and 84.49\% for EIDIG.
It is because the IDIs of NeuronFair are more diverse than those of baselines,
so it can discover more potential fairness violations and implement higher fairness improvement through retraining.
\end{itemize}

\vspace{-0.3cm}
%%%%%%%%%%%%%%%%%%%%%%%%%%%%%%%%%%%%%%%%%%%%%%%%%%%%%%%%% Answer1
\begin{center}
\fcolorbox{black}{white!20}{\parbox{0.97\linewidth}
    {
        \emph{\textbf{Answer to RQ1}}:
        NeuronFair outperforms the SOTA methods (i.e., Aequitas, SymbGen, ADF, and EIDIG) in two aspects:
        (1)~\emph{quantity} -
        it generates $\sim\!\!\times$5.84 IDIs on average compared to baselines;
        (2)~\emph{quality} -
        it searches $\sim\!\!\times$3.03 input space with more than $\sim\!\times$1.65 GSR on average compared to baselines,
        it generates IDIs that are $\sim\!\!\times$6.24 and
        $\sim\!\!\times$1.38 more diverse than Aequitas and ADF on average with $\rho_{\rm cons}<$0.02,
        it is beneficial to DNNs' fairness improvement of 87.24\% on average.
    }
}
\end{center}

\subsection{Research Questions 2}
%%%%%%%%%%%%%%%%%%%%%%%%%%%%%%%%%%%%%%%%%%%% RQ2 %%%%%%%%%%%%%%%%%%%%%%%%%%%%%%%%%%%%%%%%%%
\vspace{-0.3cm}
\begin{center}
\fcolorbox{black}{gray!20}{\parbox{0.97\linewidth}
    {
        How efficient is NeuronFair in generating IDIs?
    }
}
\end{center}

\begin{table}[t]
\caption{Time (sec) taken to generate 1,000 IDIs.}
\vspace{-0.3cm}
\label{tab:timeCost}
\resizebox{\linewidth}{!}{  \scriptsize
\begin{tabular}{llrrrrr}
\toprule
\hline
\textbf{Datasets}       & \textbf{\begin{tabular}[c]{@{}l@{}}Sen.\\ Att.\end{tabular}} & \multicolumn{1}{c}{\textbf{\begin{tabular}[c]{@{}c@{}}Aequi\\ tas\end{tabular}}} & \multicolumn{1}{c}{\textbf{\begin{tabular}[c]{@{}c@{}}Symb\\ Gen\end{tabular}}} & \multicolumn{1}{c}{\textbf{ADF}} & \multicolumn{1}{c}{\textbf{EIDIG}} & \multicolumn{1}{c}{\textbf{\begin{tabular}[c]{@{}c@{}}Neuron\\ Fair\end{tabular}}} \\ \hline
\multirow{3}{*}{Adult}  & gender                                                       & 345.68                                                                           & 1,568.20                                                                        & 298.46                           & 156.38                             & \textbf{121.56}                                                                    \\
                        & race                                                         & 1,219.35                                                                         & 5,168.24                                                                        & 268.34                           & 146.14                             & \textbf{114.25}                                                                    \\
                        & age                                                          & 484.00                                                                           & 2,431.09                                                                        & 213.76                           & 118.85                             & \textbf{105.64}                                                                    \\ \cline{2-7}
\multirow{2}{*}{GerCre} & gender                                                       & 436.00                                                                           & 2,014.68                                                                        & 488.19                           & 344.10                             & \textbf{296.46}                                                                    \\
                        & age                                                          & 531.00                                                                           & 2,834.12                                                                        & 209.44                           & 116.14                             & \textbf{103.91}                                                                    \\ \cline{2-7}
BanMar                  & age                                                          & 557.00                                                                           & 3,015.21                                                                        & 472.59                           & 246.64                             & \textbf{116.52}                                                                    \\ \cline{2-7}
COMPAS                  & race                                                         & 524.13                                                                           & 2,315.94                                                                        & 253.69                           & 199.93                             & \textbf{187.50}                                                                    \\ \cline{2-7}
MEPS                    & gender                                                       & 498.16                                                                           & 2,537.58                                                                        & 217.65                           & 182.34                             & \textbf{152.36}                                                                    \\ \hline \bottomrule
\end{tabular}
}
\vspace{-0.3cm}
\end{table}

%%%%%%%%%%%%%%%%%%%%%%%%%%%%%%%%%%%%%%%比较效率
When answering this question, we refer to the generation \textit{speed}.
The evaluation results are shown in Tab.~\ref{tab:timeCost},
where the time cost of SymbGen includes generating the explainer and constraint solving.
Here we have the following observation.

\begin{itemize}[leftmargin=9pt,topsep=0pt,partopsep=0pt]
    \item
NeuronFair generates IDIs more efficiently,
which meets the rapidity requirements of software engineering testing.
For instance,
in Tab.~\ref{tab:timeCost},
on average,
NeuronFair takes only 26.07\%, 5.47\%, 49.47\%, and 79.32\% of the time required by Aequitas, SymbGen, ADF, and EIDIG, respectively.
The outstanding performance of NeuronFair is mainly because
it uses a momentum acceleration strategy and shortens the derivation path to reduce computational complexity.
Hence, it takes less time than baselines.
\end{itemize}

\vspace{-0.4cm}
%%%%%%%%%%%%%%%%%%%%%%%%%%%%%%%%%%%%%%%% Answer2
\begin{center}
\fcolorbox{black}{white!20}{\parbox{0.97\linewidth}
    {
        \emph{\textbf{Answer to RQ2}}:
        NeuronFair is more efficient in generation \emph{speed} -
        it produces IDIs with an average speedup of 534.56\%. % than baselines.
    }
}
\end{center}

\begin{figure}[t]
\centering
\subfigure[Activated neurons by IDIs]{
    \includegraphics[width=0.46\linewidth]{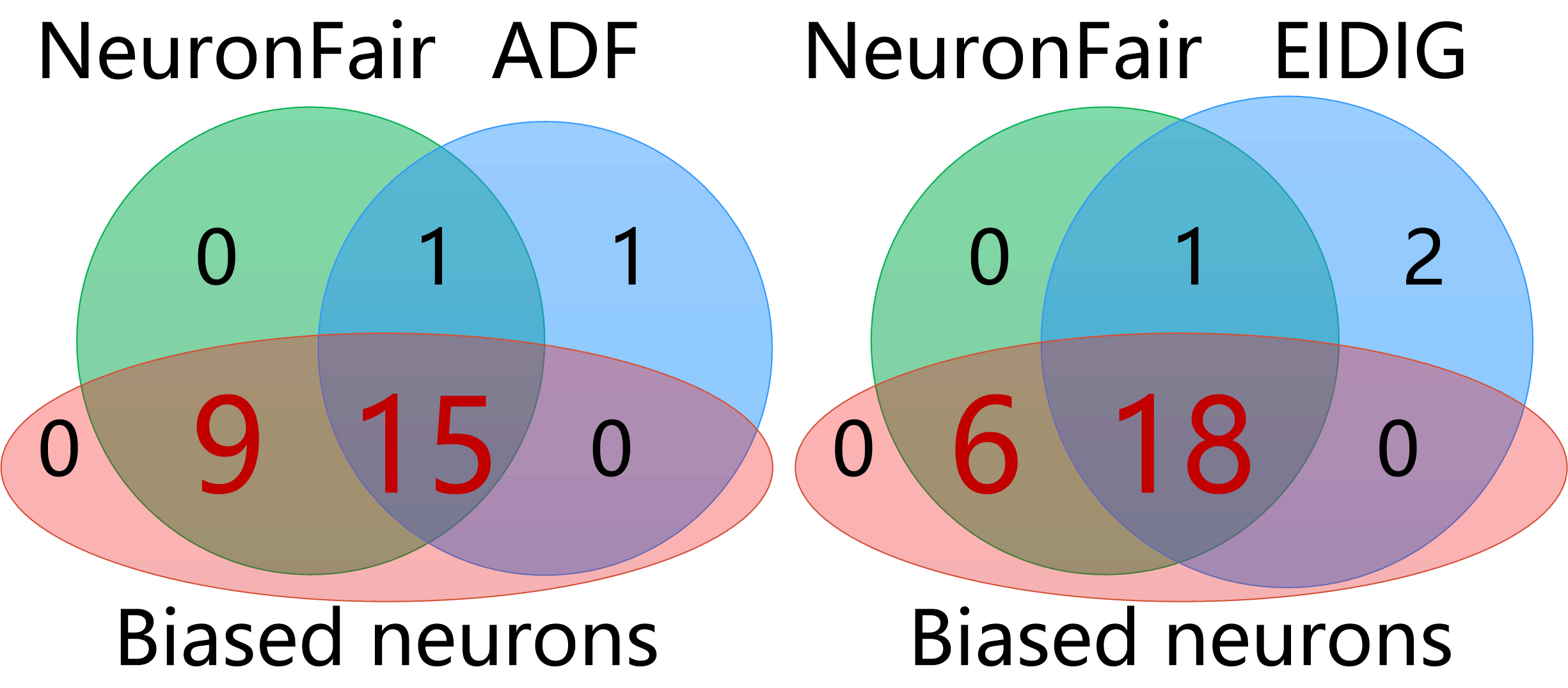}
    }
\subfigure[Activated neurons by non-IDIs]{
    \includegraphics[width=0.46\linewidth]{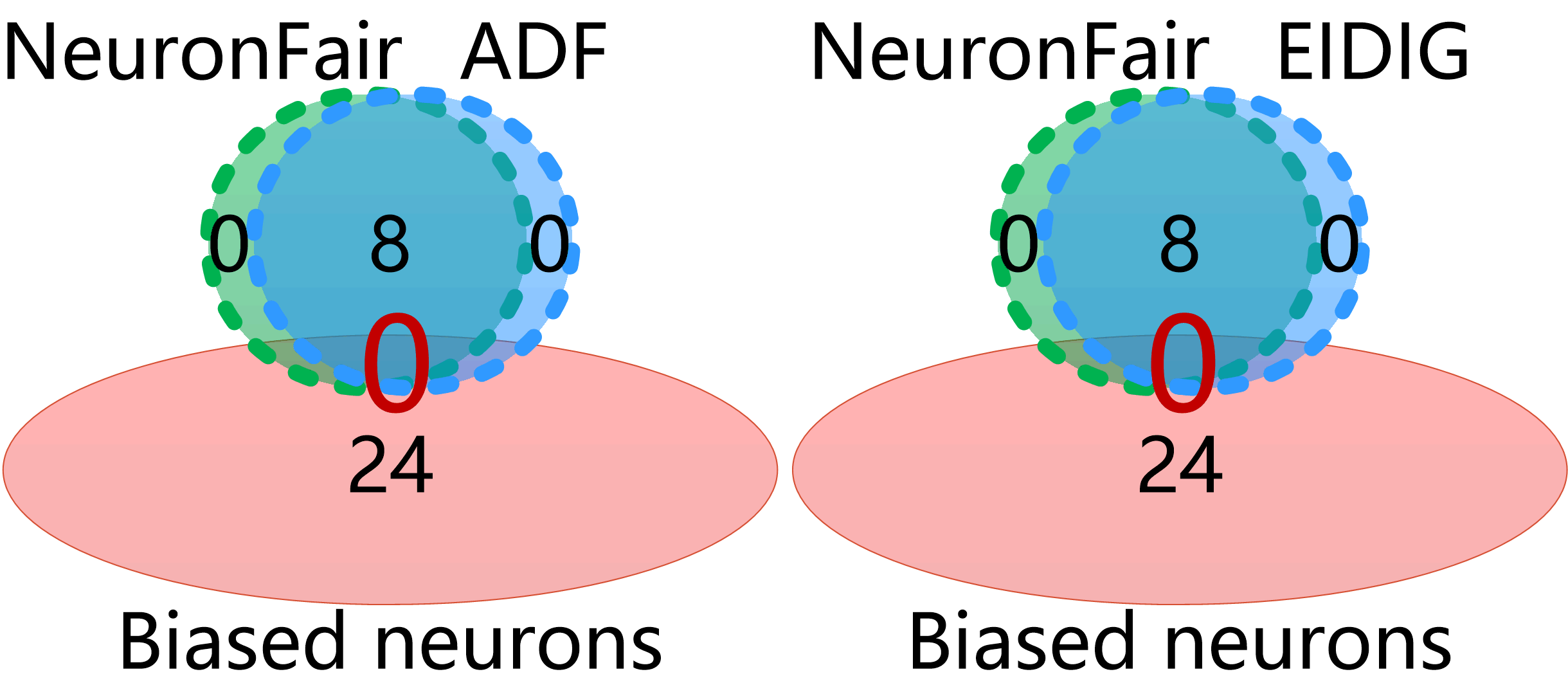}
    }
    \vspace{-0.3cm}
  \caption{
            The overlap of biased neurons and neurons activated by different instances.
        }
  \label{fig:NeuronCoverage}
  \vspace{-0.3cm}
\end{figure}

\subsection{Research Questions 3}

\vspace{-0.3cm}
\begin{center}
\fcolorbox{black}{gray!20}{\parbox{0.97\linewidth}
    {
        How to interpret NeuronFair's utility by biased neurons?
    }
}
\end{center}

When interpreting the utility,
we refer to the biased neuron \textit{coverage}.
The evaluation results are shown in Fig.~\ref{fig:NeuronCoverage}.

Implementation details for interpretation:
(1)~we conduct experiments on the Adult dataset with gender attribute for the LFC-A classifier;
(2)~we compare the interpretation results of NeuronFair with ADF and EIDIG;
(3)~for a fair interpretation,
we randomly select 10\% IDIs and 10\% non-IDIs (i.e., the generated failure instances) for each method,
and then compute the coverage of biased neurons, as shown in Fig.~\ref{fig:NeuronCoverage}.

\begin{itemize}[leftmargin=9pt,topsep=0pt,partopsep=0pt]
    \item
Biased neurons can be adopted to interpret the utility of IDIs and NeuronFair.
First,
IDIs trigger discrimination by activating biased neurons.
For instance,
the neurons activated by IDIs can cover most of the biased neurons in Fig.~\ref{fig:NeuronCoverage} (a),
while the coverage of the biased neurons by non-IDIs of different methods is 0 in Fig.~\ref{fig:NeuronCoverage} (b).
We can further interpret the utility of testing methods is related to the coverage of biased neurons,
i.e.,
NeuronFair is more effective than ADF and EIDIG because
they miss some discrimination contained in biased neurons while NeuronFair does not.
For instance,
in Fig.~\ref{fig:NeuronCoverage} (a),
the NeuronFair's IDIs activate all 24 biased neurons in the 2-nd layer of LFC-A classifier, while the neurons activated by other IDIs cannot cover all (15 for ADF and 18 for EIDIG).
\end{itemize}

\vspace{-0.4cm}
%%%%%%%%%%%%%%%%%%%%%%%%%%%%%%%%%%%%%%%% Answer3
\begin{center}
\fcolorbox{black}{white!20}{\parbox{0.97\linewidth}
    {
        \emph{\textbf{Answer to RQ3}}:
        The main reason for NeuronFair's utility is that its IDIs can activate more biased neurons.
        NeuronFair's IDIs activate 100\% biased neurons,
        while 62.5\% for ADF and 75\% for EIDIG.
    }
}
\end{center}

\subsection{Research Questions 4}
%%%%%%%%%%%%%%%%%%%%%%%%%%% RQ4 %%%%%%%%%%%%%%%%%%%%%%%%%%%%%%%%%%%%%%%%%%%%%%%%%%%%%
\begin{center}
\vspace{-0.3cm}
\fcolorbox{black}{gray!20}{\parbox{0.97\linewidth}
    {
        % \emph{\textbf{RQ1}}: How useful are the AS curve and AUC for interpreting and measuring discrimination of DNNs?
        How useful is the AUC metric for measuring DNNs' fairness?
    }
}
\end{center}

When answering this question,
we refer to the following aspects:
the \textit{consistency}, \textit{significance}, and \textit{complexity} between AUC and DM-RS.
The evaluation results on Adult and GerCre datasets with multiple sensitive attributes are shown in Tab.~\ref{tab:UtilityAUCandFairnessImprovement}.
From the results, we have the following observations.

\begin{itemize}[leftmargin=9pt,topsep=0pt,partopsep=0pt]
    \item
In all cases, AUC can correctly distinguish DNNs' fairness violations,
i.e., AUC can serve the discrimination measurement of DNN.
For instance,
in Tab.~\ref{tab:UtilityAUCandFairnessImprovement},
all of the $\rho_{s}$ values are 1.00,
indicating that the discrimination ranking results of different DNNs based on AUC are completely consistent with those based on DM-RS.
Since DNN's decision results are determined by the neurons' activation,
we speculate that the biased decisions are also caused by the neurons' activation,
i.e., neurons contain discrimination information.
Therefore, we can leverage the discrimination information in neurons to determine DNNs' fairness.
\end{itemize}

\begin{itemize}[leftmargin=9pt,topsep=0pt,partopsep=0pt]
    \item
In all cases, AUC can distinguish the different DNN's discrimination more significantly than DM-RS,
which is beneficial for a more accurate evaluation of IDIs of different testing methods.
For instance,
in Tab.~\ref{tab:UtilityAUCandFairnessImprovement},
all $\sigma$ values of AUC are higher than those of DM-RS,
and the average $\sigma$ value of AUC is 8.46 times that of DM-RS.
The outstanding performance of AUC
is mainly because we use the neurons' ActDiff to measure the discrimination,
which extracts more bias-related information from the whole DNN;
while MD-RS only uses the bias-related information from the output layer.
\end{itemize}

\begin{itemize}[leftmargin=9pt,topsep=0pt,partopsep=0pt]
    \item
The computational complexity of AUC is much lower than that of MD-RS,
which is beneficial to quickly distinguish DNNs' discrimination or the IDIs' effect.
The time frequency of AUC is ${\rm T}(N_{l})=(7+count)\times N_{l}+1$ based on \textbf{Algorithm 1}.
Thus the time complexity of AUC is $\mathcal{O}(N_{l})$,
while that of DM-RS is $\mathcal{O}(n\log n)$,
where $n$ is the instance number, $N_{l}$ is the layer number, $n>>N_{l}$.
It is mainly because AUC only conducts matrix operations,
while DM-RS requires iterative operations until convergence.
% Therefore, the calculation of AUC is simple and lightweight.
\end{itemize}

%%%%%%%%%%%%%%%%%%%%%%%%%%%%%%%%%%%%%%%% Answer4
\vspace{-0.3cm}
\begin{center}
\fcolorbox{black}{white!20}{\parbox{0.97\linewidth}
    {
        \emph{\textbf{Answer to RQ4}}:
            The AUC is useful for discrimination measurement.
            Compared to the results in Tab.~\ref{tab:UtilityAUCandFairnessImprovement},
            AUC is
            (1)~100\% \emph{consistent} with DM-RS,
            (2)~$\times$8.46 more \emph{significant} than DM-RS,
            (3)~low computational \emph{complexity} with $\mathcal{O}(N_{l})$.
    }
}
\end{center}

\begin{table}[t]
\caption{The consistency and significance between DM-RS and AUC ($\bm \rho_{s}$, $\bm \sigma$).
        Note that here we only use normal instance pairs to compute each layer's AUC,
        and select the maximum AUC as the classifier's discrimination.
        }
\vspace{-0.3cm}
\label{tab:UtilityAUCandFairnessImprovement}
\resizebox{\linewidth}{!}{ \Huge
\begin{tabular}{lllrrrrrr||lr}
\toprule[0.5mm]
\hline
\multirow{2}{*}{\textbf{Datasets}} & \multirow{2}{*}{\textbf{\begin{tabular}[c]{@{}l@{}}Sen. \\ Att.\end{tabular}}} & \multirow{2}{*}{\textbf{Metrics}} & \multicolumn{1}{c}{\multirow{2}{*}{\textbf{Before}}} & \multicolumn{5}{c||}{\textbf{After}}                                                                                                                                                                                                                                                          & \multicolumn{1}{c}{\multirow{2}{*}{\textbf{$\bm \rho_{s}$}}} & \multicolumn{1}{c}{\multirow{2}{*}{\textbf{$\bm \sigma$}}} \\ \cline{5-9}
                                   &                                                                                &                                   & \multicolumn{1}{c}{}                                 & \multicolumn{1}{c}{\begin{tabular}[c]{@{}c@{}}Aequi\\ tas\end{tabular}} & \multicolumn{1}{c}{\begin{tabular}[c]{@{}c@{}}Symb\\ Gen\end{tabular}} & \multicolumn{1}{c}{ADF} & \multicolumn{1}{c}{EIDIG} & \multicolumn{1}{c||}{\textbf{\begin{tabular}[c]{@{}c@{}}Neuron\\ Fair\end{tabular}}} & \multicolumn{1}{c}{}                                         & \multicolumn{1}{c}{}                                       \\ \hline
\multirow{6}{*}{Adult}             & \multirow{2}{*}{gender}                                                        & DM-RS                             & 2.88\%                                               & 0.45\%                                                                  & 0.44\%                                                                 & 0.26\%                  & 0.21\%                    & \textbf{0.19\%}                                                                     & \multicolumn{1}{l|}{\multirow{2}{*}{1.00}}                   & 0.0106                                                     \\
                                   &                                                                                & \textbf{AUC}                      & 0.7513                                               & 0.1492                                                                  & 0.1482                                                                 & 0.1331                  & 0.1098                    & \textbf{0.0897}                                                                     & \multicolumn{1}{l|}{}                                        & \textbf{0.2563}                                            \\ \cline{3-11}
                                   & \multirow{2}{*}{race}                                                          & DM-RS                             & 8.91\%                                               & 0.61\%                                                                  & 0.81\%                                                                 & 0.75\%                  & 0.69\%                    & \textbf{0.57\%}                                                                     & \multicolumn{1}{l|}{\multirow{2}{*}{1.00}}                   & 0.0336                                                     \\
                                   &                                                                                & \textbf{AUC}                      & 0.8045                                               & 0.1466                                                                  & 0.1795                                                                 & 0.1652                  & 0.1599                    & \textbf{0.1070}                                                                     & \multicolumn{1}{l|}{}                                        & \textbf{0.2677}                                            \\ \cline{3-11}
                                   & \multirow{2}{*}{age}                                                           & DM-RS                             & 14.56\%                                              & 4.40\%                                                                  & 4.38\%                                                                 & 4.18\%                  & 3.74\%                    & \textbf{3.30\%}                                                                     & \multicolumn{1}{l|}{\multirow{2}{*}{1.00}}                   & 0.0433                                                     \\
                                   &                                                                                & \textbf{AUC}                      & 0.8591                                               & 0.1565                                                                  & 0.1520                                                                 & 0.1482                  & 0.1347                    & \textbf{0.1082}                                                                     & \multicolumn{1}{l|}{}                                        & \textbf{0.2941}                                            \\ \cline{2-11}
\multirow{4}{*}{GerCre}            & \multirow{2}{*}{gender}                                                        & DM-RS                             & 5.16\%                                               & 0.76\%                                                                  & 0.67\%                                                                 & 0.55\%                  & 0.56\%                    & \textbf{0.49\%}                                                                     & \multicolumn{1}{l|}{\multirow{2}{*}{1.00}}                   & 0.0186                                                     \\
                                   &                                                                                & \textbf{AUC}                      & 0.6308                                               & 0.1733                                                                  & 0.1568                                                                 & 0.1422                  & 0.1524                    & \textbf{0.0960}                                                                     & \multicolumn{1}{l|}{}                                        & \textbf{0.2004}                                            \\ \cline{3-11}
                                   & \multirow{2}{*}{age}                                                           & DM-RS                             & 30.90\%                                              & 3.66\%                                                                  & 3.46\%                                                                 & 3.31\%                  & 3.21\%                    & \textbf{2.32\%}                                                                     & \multicolumn{1}{l|}{\multirow{2}{*}{1.00}}                   & 0.1132                                                     \\
                                   &                                                                                & \textbf{AUC}                      & 0.8608                                               & 0.2046                                                                  & 0.1691                                                                 & 0.1568                  & 0.1432                    & \textbf{0.1204}                                                                     & \multicolumn{1}{l|}{}                                        & \textbf{0.2879}                                            \\ \hline \bottomrule[0.5mm]
\end{tabular}
}
\vspace{-0.3cm}
\end{table}

\subsection{Research Questions 5}
%%%%%%%%%%%%%%%%%%%%%%%%%%%%%%%%%%%%%%%%%%%% RQ5 %%%%%%%%%%%%%%%%%%%%%%%%%%%%%%%%%%%%%%%%%%
\vspace{-0.3cm}
\begin{center}
\fcolorbox{black}{gray!20}{\parbox{0.97\linewidth}
    {
        % \emph{\textbf{RQ3}}: Can NeuronFair be successfully extended to the task of discriminatory image search?
        % Can NeuronFair be successfully generalized to the task of image IDI generation?
        How generic is NeuronFair for the task of image IDI generation?
    }
}
\end{center}

\begin{table}[]
\caption{`\#IDIs' and `GSR' measurements in the global phase on image datasets.}
\vspace{-0.3cm}
\label{tab:comparisonADFImage}
\resizebox{\linewidth}{!}{
\begin{tabular}{llcrcrcr}
\toprule
\hline
\multicolumn{1}{l}{\multirow{2}{*}{\textbf{Datasets}}} & \multirow{2}{*}{\textbf{Sen. Att.}} & \multicolumn{2}{c}{\textbf{ADF}}                    & \multicolumn{2}{c}{\textbf{EIDIG}}                  & \multicolumn{2}{c}{\textbf{NeuronFair}} \\ \cline{3-8}
\multicolumn{1}{l}{}                                   &                                     & \textbf{\#IDIs} & \multicolumn{1}{r|}{\textbf{GSR}} & \textbf{\#IDIs} & \multicolumn{1}{r|}{\textbf{GSR}} & \textbf{\#IDIs}    & \textbf{GSR}       \\ \hline
\multirow{2}{*}{ClbA-IN}                               & gender                              & 1,087           & \multicolumn{1}{r|}{11.58\%}      & 2,895           & \multicolumn{1}{r|}{12.50\%}      & \textbf{10,578}    & \textbf{69.90\%}   \\
                                                       & race                                & 11,908          & \multicolumn{1}{r|}{33.54\%}      & 25,180          & \multicolumn{1}{r|}{59.87\%}      & \textbf{51,529}    & \textbf{90.15\%}   \\ \cline{2-8}
\multirow{2}{*}{LFW-IN}                                & gender                              & 1,204           & \multicolumn{1}{r|}{33.20\%}      & 1,105           & \multicolumn{1}{r|}{40.10\%}      & \textbf{3,950}     & \textbf{61.40\%}   \\
                                                       & race                                & 2,269           & \multicolumn{1}{r|}{31.70\%}      & 5,304           & \multicolumn{1}{r|}{62.40\%}      & \textbf{5,457}     & \textbf{64.17\%}   \\ \hline \bottomrule
\end{tabular}
}
\vspace{-0.3cm}
\end{table}

\begin{table}[]
\caption{Fairness improvement of face detectors.}
\vspace{-0.3cm}
\label{tab:improvementImage}
\resizebox{\linewidth}{!}{\large
\begin{tabular}{llcccccccc}
\toprule
\hline
\multirow{3}{*}{\textbf{Datasets}} & \multirow{3}{*}{\textbf{\begin{tabular}[c]{@{}l@{}}Sen. \\ Att.\end{tabular}}} & \multicolumn{2}{c}{\multirow{2}{*}{\textbf{Before}}}         & \multicolumn{6}{c}{\textbf{After}}                                                                                                          \\ \cline{5-10}
                                   &                                                                                & \multicolumn{2}{c}{}                                         & \multicolumn{2}{c}{\textbf{ADF}}                & \multicolumn{2}{c}{\textbf{EIDIG}}              & \multicolumn{2}{c}{\textbf{NeuronFair}} \\ \cline{3-10}
                                   &                                                                                & \textbf{AUC} & \multicolumn{1}{c|}{\textbf{DR}}              & \textbf{AUC} & \multicolumn{1}{c|}{\textbf{DR}} & \textbf{AUC} & \multicolumn{1}{c|}{\textbf{DR}} & \textbf{AUC}       & \textbf{DR}        \\ \hline
\multirow{2}{*}{ClbA-IN}           & gender                                                                         & 0.3587       & \multicolumn{1}{c|}{\multirow{2}{*}{99.83\%}} & 0.3328       & \multicolumn{1}{c|}{97.20\%}     & 0.3091       & \multicolumn{1}{c|}{95.40\%}     & \textbf{0.1650}    & \textbf{98.40\%}   \\
                                   & race                                                                           & 0.4438       & \multicolumn{1}{c|}{}                         & 0.4045       & \multicolumn{1}{c|}{96.50\%}     & 0.3720       & \multicolumn{1}{c|}{95.50\%}     & \textbf{0.2501}    & \textbf{98.90\%}   \\ \cline{2-10}
\multirow{2}{*}{LFW-IN}            & gender                                                                         & 0.3910       & \multicolumn{1}{c|}{\multirow{2}{*}{99.56\%}} & 0.3524       & \multicolumn{1}{c|}{95.30\%}     & 0.3678       & \multicolumn{1}{c|}{92.30\%}     & \textbf{0.1091}    & \textbf{98.90\%}   \\
                                   & race                                                                           & 0.4251       & \multicolumn{1}{c|}{}                         & 0.3984       & \multicolumn{1}{c|}{98.10\%}     & 0.3933       & \multicolumn{1}{c|}{96.40\%}     & \textbf{0.2240}    & \textbf{99.10\%}   \\ \hline \bottomrule
\end{tabular}
}
\vspace{-0.3cm}
\end{table}

When reporting the results, we focus on two aspects:
generation \textit{quantity} and \textit{quality}.

Implementation details for NeuronFair generalized on image data:
(1)~we only perform comparisons with ADF and EIDIG at global phase,
because the effect of ADF and EIDIG on DNNs is much better than that of Aequitas and SymbGen;
(2)~we remove the $\rm KMeans(\cdot)$ operation,
set $step_{-}size_{g}$=0.15 for image,
% $\rm NUM(NdIs)$=$num_{g}$,
and all face images are used as input;
(3)~we retrain the face detector with all image IDIs of each method and measure its fairness improvement by AUC,
(4)~we measure the bias perturbation $\Delta_{bias}$ and sensitive attribute perturbation $\Delta_{senatt}$ by $L_{2}$-norm.

\textbf{Generation Quantity}.
The evaluation results are shown in
Tab.~\ref{tab:comparisonADFImage} measured by the IDIs number in \textit{global} phase.
From the results, we have the following observation.

\begin{itemize}[leftmargin=9pt,topsep=0pt,partopsep=0pt]
    \item
In all cases, NeuronFair can obtain more IDIs than ADF and EIDIG,
especially for the discrimination against race attribute.
For instance,
in Tab.~\ref{tab:comparisonADFImage},
on average,
NeuronFair generates 4.34 times and 2.07 times IDIs of ADF and EIDIG, respectively.
The outstanding performance of NeuronFair is because it adopts dynamic loss to expand the valid input space
while ADF and EIDIG do not consider the dynamics of search.
Meanwhile,
the number of IDIs generated by NeuronFair for race is 3.91 times that for gender.
We speculate the reason is that
the pixel information related to race is mainly skin color (i.e., light \& dark, or black \& white),
while the pixel information related to gender is more diverse (such as hair, makeup, face shape, etc.).
Therefore, the image IDIs generation for race is easier through manipulating skin color.
\end{itemize}

\textbf{Generation Quality}.
The evaluation results are shown in
Tabs.~\ref{tab:comparisonADFImage} and \ref{tab:improvementImage},
including three scenarios:
the generation success rate (\textit{GSR}),
the fairness improvement (\textit{AUC}),
and the detection rate (\textit{DR}).

\begin{itemize}[leftmargin=9pt,topsep=0pt,partopsep=0pt]
    \item
Among image data,
image IDIs of NeuronFair are of higher quality than those of ADF and EIDIG,
which can be applied to retrain face detectors and contribute to their fairness improvement in face detection scenarios.
For instance,
in Tab~\ref{tab:comparisonADFImage},
on average,
the GSR value of NeuronFair is 2.60 times and 1.63 times that of ADF and EIDIG, respectively.
It is because NeuronFair reduces the probability of gradient vanishing,
which in turn improves the probability of non-duplicate IDIs generation guided by the gradient.
Hence, all GSR values of NeuronFair are higher than those of baselines.
Meanwhile,
the valid input space of NeuronFair is 1.67 times and 1.27 times that of ADF and EIDIG, respectively.
Since the probability of falling into a local optimum is reduced by dynamically combining biased neurons,
we can perform valid searches in limited instance space.
\end{itemize}

\begin{itemize}[leftmargin=9pt,topsep=0pt,partopsep=0pt]
    \item
NeuronFair contributes more to the fairness improvement of the face detector,
i.e.,
its generalization on image data is better than that of ADF and EIDIG.
For instance,
in Tab.~\ref{tab:improvementImage},
on average,
the discrimination of detectors retrained with IDIs of NeuronFair dropped by 53.77\%,
while the AUC values of ADF and EIDIG only dropped by 8.06\% and 10.90\%, respectively.
We speculate that the valid input space of NeuronFair is larger,
so its IDIs can find potential discrimination that other methods' IDIs cannot.
Then improve the detector's fairness through retraining.
\end{itemize}

\begin{itemize}[leftmargin=9pt,topsep=0pt,partopsep=0pt]
    \item
NeuronFair hardly affects the detector's DR values while improving its fairness.
For instance,
in Tab.~\ref{tab:improvementImage},
on average,
the DR value of detectors retrained with NeuronFair's IDIs only dropped by 0.87\%,
while that of ADF and EIDIG dropped by 2.92\% and 4.80\%, respectively.
We compare the $L_{2}$ norm of $\Delta_{bias}$ and $\Delta_{senatt}$ generated by different methods,
and find that $\Delta_{bias}$ of NeuronFair is much lower than that of ADF and EIDIG. Therefore, NeuronFair can not only improve the detector's fairness but also maintain its detection performance.
\end{itemize}

%%%%%%%%%%%%%%%%%%%%%%%%%%%%%%%%%%%%%%%% Answer5
\vspace{-0.4cm}
\begin{center}
\fcolorbox{black}{white!20}{\parbox{0.97\linewidth}
    {
    \emph{\textbf{Answer to RQ5}}:
        The generalization performance of NeuronFair on the image dataset is better than the SOTA methods (i.e., ADF and EIDIG) in two aspects:
        (1)~\emph{quantity} -
        it generates $\sim\!\!\times$4.34 and $\sim\!\!\times$2.07 image IDIs on average compared to ADF and EIDIG, respectively;
        (2)~\emph{quality} -
        it searches $\sim\!\!\times$1.47 input space with more than $\sim\!\times$2.11 GSR on average,
        it is beneficial to detectors' fairness improvement of 53.77\% on average but hardly affects their detection performance.
        Thus, NeuronFair shows better generalization performance than ADF and EIDIG.
    }
}
\end{center}

\section{Threats to Validity}

\quad\textbf{Correlation between attributes}.
The attributes of unstructured data are not as clear as structured data,
so we provide a generalization framework that can modify sensitive attributes.
However, there is a correlation between attributes,
i.e., after the perturbation for one sensitive attribute is added,
another attribute may also be changed.
Since the transferability of perturbation is not robust.
the slight attribute change will not affect our IDI generation.

\textbf{Sensitive attributes}.
We consider only one sensitive attribute at a time for our experiments.
However, considering multiple protected attributes will not hamper the effectiveness or generalization offered by our novel testing technique,
but will certainly lead to an increase in execution time.
This increase is attributed towards the fact that the algorithm in such a case,
needs to consider all the possible combinations of their unique values.

\textbf{Access to DNNs}.
NeuronFair is white-box testing that generates IDIs based on the biased neurons inside DNNs,
which means it requires accessing to DNNs.
It is widely accepted that DNN testing could have full knowledge of the target model in software engineering.

\section{Related Works\label{RelatedWorks}}

\quad\textbf{Fairness Testing}.
Based on the software engineering point of view,
several works on testing the fairness of traditional ML models are proposed~\cite{Tramer2017FairTest,Adebayo2016FairML,Udeshi2018Automated,Galhotra2017Fairness,Aggarwal2018Automated,Zhang2021Ignorance}.
To uncover their fairness violations,
Galhotra et al.~\cite{Galhotra2017Fairness} firstly proposed Themis, a fairness testing method for software,
which measures the discrimination in software through counting the frequency of IDIs in the input space.
However, its efficiency for IDIs generation is unsatisfactory.
To improve the generation speed of Themis,
Udeshi et al.~\cite{Udeshi2018Automated} proposed a faster generation algorithm, Aequitas,
which uncovers fairness violations by probabilistic search over the input space.
Aequitas adopts a two-phase operation in which the IDIs generated globally are used as seeds for the local generation.
% Its efficiency has been greatly improved.
However, Aequitas uses a global sampling distribution for all the inputs,
which leads to the limitation that it can only search in narrow input space and easily falls into the local optimum.
Thus, Aequitas's IDIs lack diversity.
To further improve the instance diversity,
Agarwal et al.~\cite{Aggarwal2018Automated} designed a new testing method, SymbGen,
which combines the symbolic execution along with the local interpretation for the generation of effective instances.
SymbGen constructs the local explainer of the complex model at first
and then searches for IDIs based on the fitted decision boundary.
Therefore, its instance effectiveness almost depends on the performance of the explainer.

The above-mentioned methods mainly deal with traditional ML models,
which cannot directly be applied to deal with DNNs.
Recently, several methods have been proposed specifically for DNNs.
For instance,
Zhang et al.~\cite{Zhang2020White} first proposed a fairness testing method specifically for DNNs, ADF,
which guides the search direction through gradients.
The authors proved that its effectiveness and efficiency of IDIs generation for DNNs are greatly improved based on the guidance of gradients.
Based on the ADF~\cite{Zhang2020White},
Zhang et al.~\cite{Zhang2021Efficient} designed a framework EIDIG for discovering individual fairness violations,
which adopts prior information to accelerate the convergence of iterative optimization.
However, there is still a problem of gradient vanishing, which may lead to local optimization.

\textbf{Neuron-based DNN Interpretation}.
Kim et al.~\cite{Kim2018Interpretability} first introduced concept activation vectors,
which provide an interpretation of a DNN's internal state (i.e., the activation output in the hidden layer).
They viewed the high-dimensional internal state of a DNN as an aid,
and interpreted which concept is important to the classification result.
Inspired by the concept activation vectors,
Du et al.~\cite{Du2019Fairness} suggested that interpretability can serve as a useful ingredient to diagnose the reasons that lead to algorithmic discrimination.
The above methods study the activation output of one hidden layer,
while Liu et al.~\cite{Liu2019ABS} studied the activation state of a single neuron.
They observed that the neuron activation is related to the DNNs' robustness,
and used the abnormal activation of a single neuron to detect backdoor attacks.
These methods leverage the internal state to interpret DNNs' classification performance and robustness,
which inspires us to use it to interpret DNNs' biased decision.

\section{Conclusions\label{Conclusions}}

We propose an interpretable white-box fairness testing method, NeuronFair, to efficiently generate IDIs for DNNs based on biased neurons.
Our method provides discrimination interpretation and IDI generation for different data forms.
In the discrimination interpretation,
AS curve and AUC measurement are designed to qualitatively and quantitatively interpret the severity of discrimination in each layer of DNNs, respectively.
In the IDI generation,
a global phase and a local phase collaborate to systematically search the input space for IDIs with the guidance of momentum acceleration and dynamic loss.
Further, NeuronFair can process not only structured data but also unstructured data, e.g., image, text, etc.
We compare NeuronFair with four SOTA methods in 5 structured datasets and 2 face image datasets against 7 DNNs,
the results show that NeuronFair has significantly better performance in terms of
interpretability, generation effectiveness, and data generalization.

\section*{Acknowledgment}
This research was supported by
the National Natural Science Foundation of China (Nos. 62072406, 62102359),
the Key R\&D Projects in Zhejiang Province (Nos. 2021C01117 and 2022C01018),
the ``Ten Thousand Talents Program'' in Zhejiang Province (2020R52011).

\bibliographystyle{ACM-Reference-Format}
\bibliography{myref}

\end{document}